\documentclass{article}

\usepackage{arxiv}

\usepackage[utf8]{inputenc} 
\usepackage[T1]{fontenc}    
\usepackage{hyperref}       
\usepackage{url}            
\usepackage{booktabs}       
\usepackage{amsfonts}       
\usepackage{nicefrac}       
\usepackage{microtype}      
\usepackage{graphicx}
\usepackage{adjustbox}
\graphicspath{ {./images/} }
\usepackage{booktabs}
\usepackage{pifont}
\usepackage{xcolor}
\usepackage{caption}
\usepackage{amssymb}
\usepackage{array}
\usepackage[most]{tcolorbox}
\usepackage{biblatex}
\usepackage{amsmath}
\usepackage{amsthm}
\usepackage{subcaption}
\usepackage{algorithm}
\usepackage{algpseudocode}
\usepackage{float}
\usepackage{flafter}
\usepackage{placeins}

\newcommand{\good}{\textcolor{green!60!black}{\ding{51}}}     
\newcommand{\reject}{\textcolor{red!70!black}{\ding{55}}}     
\definecolor{wrongred}{HTML}{A33A3A}
\definecolor{castgreen}{HTML}{237A57}
\newtcolorbox{modeloutput}[4][]{
  enhanced,
  breakable,
  colback=white,
  colframe=#4!70!black,
  colbacktitle=#4!10,
  coltitle=black,
  fonttitle=\bfseries\small,
  fontupper=\small,
  title={#2\hfill{\normalfont #3}},
  boxrule=0.45pt,
  arc=2pt,
  left=7pt,
  right=7pt,
  top=7pt,
  bottom=7pt,
  before skip=8pt,
  after skip=8pt,
  #1
}

\title{CAST: Non-Privileged Clipped Asymmetric Self-Teaching with Advantage Flipping for GRPO}

\author{
Yang Li$^1$ \quad
Gongle Xue$^2$ \quad
Yijia Guo$^3$ \quad
Yuheng Yuan$^3$ \quad
Liwen Hu$^3$ \quad
Lei Ma$^{4\dagger}$ \\
$^1$ School of Software and Microelectronics, Peking University, Beijing \\
$^2$ School of Artificial Intelligence, Peking University, Beijing \\
$^3$ School of Computer Science, Peking University, Beijing \\
$^4$ School of Future Technology, Peking University, Beijing \\
{\small \texttt{ly0376@stu.pku.edu.cn} \quad \texttt{lei.ma@pku.edu.cn}} \\
$^\dagger$ Corresponding author.
}

\begin{document}
\maketitle
\begin{abstract}
Reinforcement learning with verifiable rewards (RLVR), especially Group Relative Policy Optimization (GRPO), has been widely used to improve reasoning in large language models. However, outcome-level rewards provide only sparse supervision, and group-relative advantages vanish when all sampled trajectories for a prompt are either correct or incorrect. On-Policy Self-Distillation (OPSD) offers dense token-level guidance, but its token preferences are not necessarily aligned with trajectory correctness; empirical diagnostics show that OPSD signals behave differently on correct and incorrect rollouts, with teacher-positive and teacher-negative gap signals exhibiting different noise profiles. These diagnostics are conducted under an OPSD-style privileged teacher context for analysis only, whereas CAST training uses answer-free self-teacher scoring.

Motivated by these observations, this work proposes \textbf{CAST} (Non-Privileged \textbf{C}lipped \textbf{A}symmetric \textbf{S}elf-\textbf{T}eaching with Advantage Flipping for GRPO), an answer-free self-distillation method for GRPO-style RLVR. CAST keeps the verifier-grounded GRPO objective, but uses a stop-gradient self-teacher to shape token-level advantages according to trajectory correctness. Unlike prior self-distilled RLVR methods, CAST does not require reference-solution-conditioned teacher scoring, keeps the self-teacher log-probability gap active throughout training, and applies bidirectional local advantage sign reversal: teacher-negative tokens in correct trajectories can receive negative token-level advantages, while teacher-positive tokens in incorrect trajectories can receive bounded positive local advantages. For zero-variance all-correct and all-wrong groups, CAST assigns bounded sign-constrained base advantages, so these otherwise zero-gradient groups can contribute verifier-signed token feedback. Experiments on mathematical reasoning show that CAST improves RLVR training while retaining a lightweight, verifier-grounded trajectory-level objective.

\end{abstract}

\begin{figure}[h]
    \centering
    \includegraphics[width=0.8\linewidth]{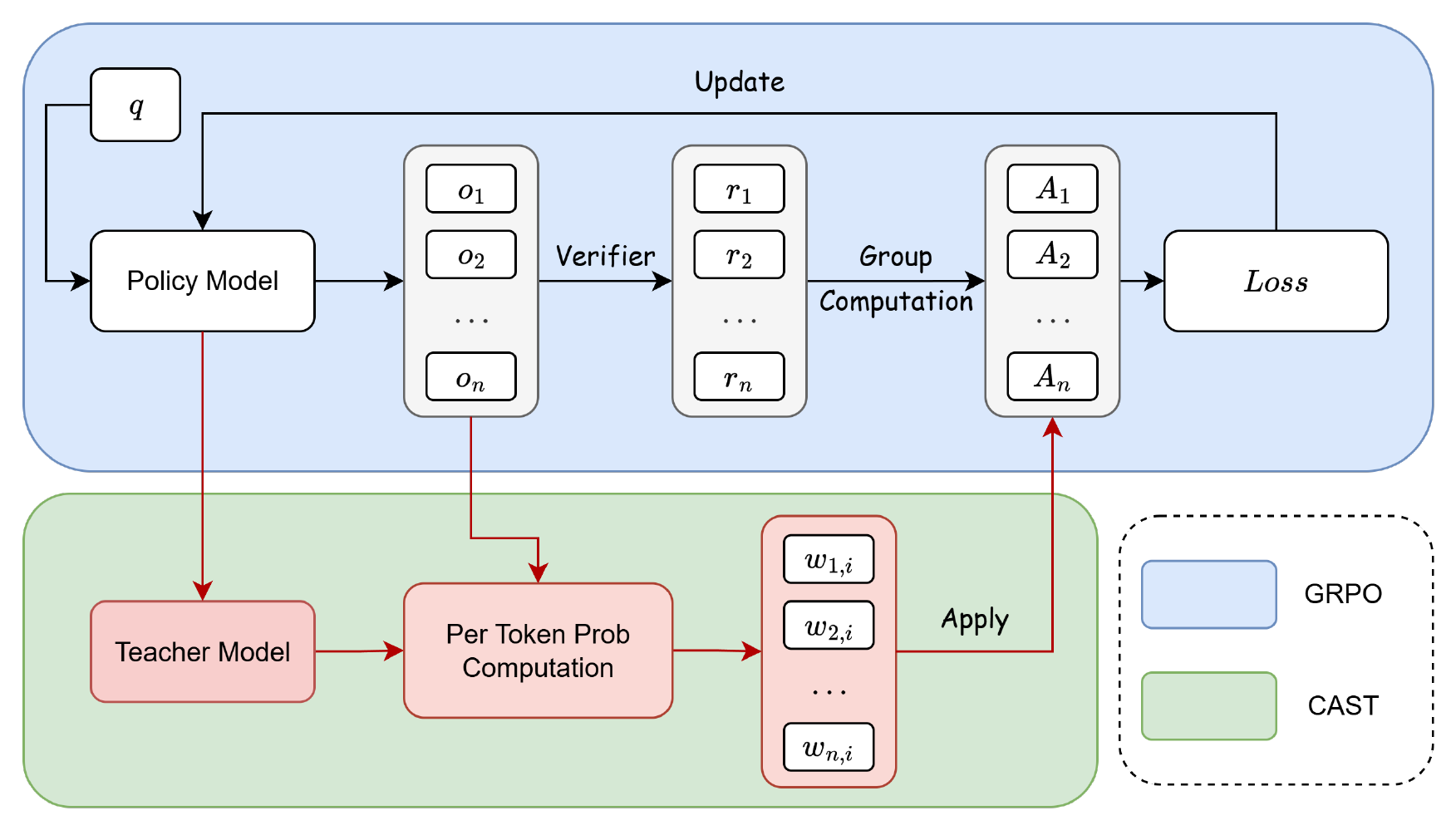}
    \caption{Overview of GRPO and CAST.}
    \label{fig:pipeline}
\end{figure}

\section{Introduction}

Reinforcement learning with verifiable rewards (RLVR) has become a widely used paradigm for improving reasoning in large language models~\cite{deepseekv3,qwen3,kimi_vl,deepseekr1,rlvr_beyond_math}. Group Relative Policy Optimization (GRPO)~\cite{grpo} samples multiple trajectories per prompt and estimates group-relative advantages from verifier outcomes, offering a scalable alternative to critic-based policy optimization~\cite{self_consistency,gsm8k_verifiers,math_dataset}.

Despite its success, GRPO provides only sparse, outcome-level supervision~\cite{process_supervision,dapo,rlsd,sdpo}: it says little about which tokens help or hurt, and the group-relative advantage vanishes when all rollouts for a prompt are correct or all are incorrect~\cite{dapo,zero_variance_prompts}. These zero-variance groups still contain useful consolidation or failure signals, motivating dense token-level guidance that remains verifier-grounded.

On-policy self-distillation~\cite{opd,opd_iclr,opsd,star,rest_lm} supplies such guidance, and recent RLVR variants~\cite{rlsd,rlrt} already exploit teacher--student gaps beyond sparse rewards. However, three issues remain: gaps are not interpreted differently on correct versus incorrect trajectories, teacher scoring often uses answer-privileged contexts, and all-correct/all-wrong groups are typically underused once group-relative advantages collapse~\cite{opsd_compresses}. Section~\ref{sec:method_motivation} studies gap structure under an OPSD-style privileged diagnostic; CAST training uses answer-free self-teacher scoring only.

We propose \textbf{CAST} (Non-Privileged \textbf{C}lipped \textbf{A}symmetric \textbf{S}elf-\textbf{T}eaching with Advantage Flipping for GRPO). CAST keeps the verifier-grounded GRPO objective, shapes token advantages with detached rollout/reference teacher--student gaps, applies bidirectional local sign reversal, and assigns bounded base advantages to zero-variance groups.

The contributions are:
\begin{itemize}
    \item Teacher--student gaps are analyzed under mathematical RLVR, showing different behavior across correctness and gap sign (Section~\ref{sec:method_motivation}).
    \item \textbf{CAST} is introduced: answer-free self-teacher scoring, asymmetric clipping, bidirectional advantage flipping, and a bounded zero-variance branch within the standard GRPO update.
    \item Experiments on Qwen3-1.7B, Qwen3-4B, and Qwen3-8B show strong overall gains under a matched 300-step LoRA budget; appendix analyses report output length and generation-token budget sensitivity.
\end{itemize}

\section{Preliminaries}

Consider mathematical reasoning tasks with verifiable final answers~\cite{gsm8k_verifiers,math_dataset,minerva}. Given a prompt $x$, a rollout/reference policy $\pi_{\theta_{\mathrm{old}}}$ samples $G$ on-policy trajectories $\{y_i\}_{i=1}^{G}$, where $y_i=(y_{i,1},\ldots,y_{i,T_i})$. A verifier assigns a binary reward $r_i\in\{0,1\}$ according to final-answer correctness. During the subsequent policy update, $\pi_\theta$ denotes the online policy being optimized and $\pi_{\theta_{\mathrm{old}}}$ denotes the detached policy used to sample the trajectories and compute reference log-probabilities. Under answer-free self-teacher scoring, the ground-truth answer is used only for verification.

GRPO~\cite{grpo} estimates a sequence-level advantage by normalizing rewards within each sampled group:
\begin{equation}
A_i =
\frac{r_i-\bar r}{\sigma_r+\epsilon_{\mathrm{std}}},
\qquad
\bar r=\frac{1}{G}\sum_{j=1}^{G}r_j .
\end{equation}
The advantage is then assigned to all valid tokens and optimized with a PPO-style clipped policy-gradient objective~\cite{ppo}. GRPO is straightforward and verifier-grounded, but when all sampled trajectories for a prompt are correct or all are incorrect, the reward variance collapses and the normalized advantage becomes zero or uninformative. These cases are denoted \textsc{AllCorrect} and \textsc{AllWrong}; groups containing both outcomes are denoted \textsc{Mixed}.

CAST augments this sparse signal with a stop-gradient self-teacher. For a teacher context $c_i$, define the token-level teacher--student gap
\begin{equation}
g_{i,t}
=
\log \pi_{\phi}(y_{i,t}\mid c_i,y_{i,<t})
-
\log \pi_{\theta_{\mathrm{old}}}(y_{i,t}\mid x,y_{i,<t}) .
\end{equation}
A token is \emph{teacher-positive} when the teacher--student gap satisfies $g_{i,t}>0$, meaning the self-teacher assigns higher log-probability to the sampled token than the detached rollout/reference policy; it is \emph{teacher-negative} when $g_{i,t}<0$. Throughout this paper, the ``student'' in teacher--student gap refers to this detached rollout/reference policy \(\pi_{\theta_{\mathrm{old}}}\), rather than the online policy after the current update begins. Prior self-distilled RLVR methods usually use such gaps to modulate the magnitude of a verifier-determined update, often without changing its sign. CAST instead interprets the gap through trajectory-level correctness, because teacher-positive tokens in incorrect trajectories and teacher-negative tokens in correct trajectories should not be treated symmetrically.

In the default setting, CAST uses answer-free self-teacher scoring:
\begin{equation}
g^{\mathrm{AF}}_{i,t}
=
\log \pi_{\phi}(y_{i,t}\mid x,y_{i,<t})
-
\log \pi_{\theta_{\mathrm{old}}}(y_{i,t}\mid x,y_{i,<t}) .
\end{equation}
Thus, the teacher observes only the prompt and generated prefix; the ground-truth answer is reserved for the verifier. Additional background on OPD/OPSD, DAPO-style GRPO variants, RLSD, and RLRT is given in Appendix~\ref{app:related_work_details}.

All token-gap quantities used to construct CAST advantages are computed on sampled tokens before the policy-gradient update and are treated as detached coefficients inside the loss. Gradients therefore flow through the policy-ratio term, not through the teacher--student gap or the clipping operations used to build \(\tilde A_{i,t}\).

\section{Method}

\subsection{Motivation: Correctness-Aware Self-Teacher Shaping}
\label{sec:method_motivation}

Self-distilled RLVR methods use On-Policy Self-Distillation (OPSD) to provide dense token-level guidance. Given a student rollout, OPSD compares the probability assigned by a stop-gradient self-teacher and the current student to each generated token. Standard OPSD treats this teacher--student gap as a generic token preference: tokens with higher teacher log-probability are encouraged, while tokens with lower teacher log-probability are suppressed.

In RLVR, however, each trajectory is also labeled by verifier correctness. This trajectory-level label changes how the teacher signal should be interpreted. A teacher-positive token inside an incorrect trajectory should not receive the same credit as one inside a correct trajectory, and a correct trajectory may still contain locally undesirable tokens such as redundant answer templates or termination markers. Uniformly reinforcing every token in a successful rollout can therefore consolidate spurious reasoning; conversely, local suppression may benefit some teacher-negative tokens when the gap indicates low support, while final advantage clipping limits over-suppression risk.

To make this concrete, an OPSD token-shift diagnostic on 2{,}048 rollouts from 256 prompts (926 correct and 1{,}122 incorrect trajectories) compares four strata formed by trajectory correctness and the sign of the per-token gap \(\Delta_t=\log p_{\mathrm{teacher}}(y_t)-\log p_{\mathrm{student}}(y_t)\). The diagnostic studies gap structure under an OPSD-style privileged teacher context; all CAST training runs below use answer-free self-teacher scoring (Appendix~\ref{app:implementation_details}). Appendix~\ref{app:opsd_motivation_diagnostic} defines the gap-weighted Top-$K$ stability metrics and reports token-level interpretations.

\begin{table}[h]
\centering
\setlength{\tabcolsep}{5pt}
\begin{tabular}{lccc}
\toprule
Stratum & Mean \( |\Delta| \)@10 & Var@10 & CV@10 \(\downarrow\) \\
\midrule
\(\mathcal{B}_{\mathrm{corr}}^{+}\) & 0.433 & 2.038 & 4.133 \\
\(\mathcal{B}_{\mathrm{corr}}^{-}\) & \textbf{9.782} & \textbf{10.416} & \textbf{1.436} \\
\(\mathcal{B}_{\mathrm{inc}}^{+}\) & 0.509 & 2.506 & 3.841 \\
\(\mathcal{B}_{\mathrm{inc}}^{-}\) & 8.306 & 8.341 & 1.896 \\
\bottomrule
\end{tabular}
\caption{Top-10 gap stability across correctness--sign strata in the OPSD token-shift diagnostic ($K{=}10$). Lower CV@10 indicates more repeatable gap mass on the dominant token types.}
\label{tab:opsd_topk_stability}
\end{table}

Table~\ref{tab:opsd_topk_stability} supports a correctness-aware reading of self-teacher gaps. Teacher-negative tokens are much larger in magnitude than teacher-positive tokens, so raw variance alone is scale-dominated; after gap-weighted normalization, teacher-negative strata have the lowest CV@10 and the most interpretable dominant tokens (e.g., termination and answer-finalization markers). Teacher-positive tokens in incorrect trajectories remain comparatively noisy. CAST therefore uses verifier correctness to set trajectory-level direction, treats teacher-negative tokens as a strong local suppression term, and allows only bounded positive credit from teacher-positive tokens in incorrect trajectories rather than trajectory-level imitation. Representative full-token maps are given in Appendix~\ref{app:token_shift_visualization}.

\subsection{Non-Privileged Self-Teacher Scoring}

Under answer-free self-teacher scoring, CAST computes the self-teacher gap for each token $y_{i,t}$ in trajectory $y_i$ as
\begin{equation}
    g_{i,t}
    =
    \log \pi_{\phi}(y_{i,t}\mid x,y_{i,<t})
    -
    \log \pi_{\theta_{\mathrm{old}}}(y_{i,t}\mid x,y_{i,<t}),
\end{equation}
where $x$ is the prompt, $\pi_{\theta_{\mathrm{old}}}$ is the detached rollout/reference student policy, and $\pi_\phi$ is a stop-gradient self-teacher. The teacher receives the same problem context and generated prefix as the student. The ground-truth answer and reference solution are used only by the verifier to compute the binary outcome reward, not in the self-teacher scoring context. Reference-solution-conditioned teacher scoring is used only as an ablation in Table~\ref{tab:ablation_components}; implementation details are given in Appendix~\ref{app:implementation_details}.

\subsection{Trajectory-Level Base Advantage}

For each prompt $x$, CAST samples $G$ on-policy trajectories
$\{y_i\}_{i=1}^{G}$ and obtains binary verifier rewards
$r_i\in\{0,1\}$. Let
\[
\tau \in \{\textsc{Mixed},\textsc{AllCorrect},\textsc{AllWrong}\}
\]
denote the group type. In mixed groups, CAST uses the standard group-relative GRPO advantage:
\begin{equation}
    A_i
    =
    \frac{r_i-\bar r}{\sigma_r+\epsilon},
    \qquad
    \bar r=\frac{1}{G}\sum_{j=1}^{G}r_j.
\end{equation}

When all sampled trajectories are correct or all are wrong, the reward variance collapses and GRPO provides no useful group-relative signal. CAST instead assigns a bounded correctness-aware base advantage:
\begin{equation}
B_i =
    \begin{cases}
        A_i, & \tau=\textsc{Mixed},\\
        +b_{\mathrm{correct}}, & \tau=\textsc{AllCorrect},\\
        -b_{\mathrm{wrong}}, & \tau=\textsc{AllWrong}.
    \end{cases}
\end{equation}
By default, $b_{\mathrm{correct}}=b_{\mathrm{wrong}}=1$.
This branch should be interpreted as a bounded verifier-signed shaping heuristic rather than an unbiased group-relative advantage estimator. We choose unit base values because final token advantages are clipped to $[-1.2,1.2]$, which keeps the effective scale comparable to clipped mixed-group updates.

This design prevents uniform-reward groups from receiving vanishing updates. In standard GRPO, all-correct and all-wrong groups have vanishing normalized advantages; in CAST, they still receive sign-constrained base advantages and can therefore contribute bounded token-level credit assignment. Consequently, CAST does not collapse to standard GRPO solely because a rollout group has zero reward variance.

\subsection{Correctness-Aware Token Weighting}

CAST converts the teacher--student gap into a signed token weight:
\begin{equation}
z_{i,t}
=
\exp\left(\operatorname{sign}(B_i)g_{i,t}\right).
\end{equation}
When $B_i>0$, a teacher-positive gap ($g_{i,t}>0$) means the teacher assigns higher log-probability to the sampled token than the student, so $z_{i,t}>1$ strengthens the positive update. When $B_i<0$, the sign is reversed in this sign-preserving weighting stage: a teacher-positive token in a negative-base trajectory receives weaker negative pressure, while a teacher-negative token receives stronger negative pressure. In our binary-reward mixed groups, $B_i=0$ does not occur unless the group is uniform, and uniform groups are handled separately by fixed-sign base advantages; for completeness, a zero-base token can be masked or assigned zero advantage. The next stage then allows sufficiently teacher-positive negative-base tokens to undergo sign reversal into bounded positive local credit.

CAST applies asymmetric clipping:
\begin{equation}
    w_{i,t}
    =
    \begin{cases}
        \mathrm{clip}(z_{i,t}, \ell_+, u_+), & B_i>0,\\
        \mathrm{clip}(z_{i,t}, \ell_-, u_-), & B_i<0.
    \end{cases}
\end{equation}
By default,
\[
(\ell_+,u_+)=(0.8,1.05),
\qquad
(\ell_-,u_-)=(0.95,1.2).
\]
This biases shaping toward stronger teacher-negative suppression than teacher-positive amplification: positive-base updates can be weakened more than they can be amplified, and negative-base updates can be strengthened more than they can be weakened. This reflects the empirical observation that teacher-positive signals are often noisier, while teacher-negative signals provide a more stable suppression term.
The incorrect-trajectory sign-reversal rule below reuses the smaller positive cap \(u_+\), so teacher-positive tokens inside negative-base trajectories receive only bounded local positive credit.

\subsection{CAST Token Advantage}

CAST first forms a sign-preserving shaped advantage:
\begin{equation}
\begin{aligned}
    \hat{A}_{i,t} &= B_i\left(1+\lambda(w_{i,t}-1)\right),
\end{aligned}
\end{equation}
where $\lambda$ controls the strength of self-teacher shaping.

CAST then applies bidirectional local advantage sign reversal. For positive-base trajectories, teacher-negative tokens can receive negative token-level advantages. If $B_i>0$ and $g_{i,t}<0$, the teacher assigns lower log-probability to the sampled token than the student. CAST treats this as a local suppression term and forms a sign-reversed candidate advantage:
\begin{equation}
\begin{aligned}
    A^{\mathrm{down}}_{i,t} &= -\lvert B_i\rvert\bigl(1+\lambda(w^{-}_{i,t}-1)\bigr),
    && B_i>0,\ g_{i,t}<0,
\end{aligned}
\end{equation}
where
\begin{equation}
w^{-}_{i,t}
=
\mathrm{clip}
\left(
\exp(-g_{i,t}),
\max(1,\ell_-),
u_-
\right).
\end{equation}
For negative-base trajectories, teacher-positive tokens can receive bounded positive token-level advantages. If $B_i<0$ and $g_{i,t}>0$, CAST treats the token as locally useful partial reasoning inside an otherwise incorrect trajectory and forms a sign-reversed positive candidate advantage:
\begin{equation}
\begin{aligned}
    A^{\mathrm{up}}_{i,t} &= \lvert B_i\rvert\bigl(1+\lambda(w^{+}_{i,t}-1)\bigr),
    && B_i<0,\ g_{i,t}>0,
\end{aligned}
\end{equation}
where
\begin{equation}
w^{+}_{i,t}
=
\mathrm{clip}
\left(
\exp(g_{i,t}),
\max(1,\ell_+),
u_+
\right).
\end{equation}
The unclipped token advantage is then defined by the following case distinction:
\begin{equation}
    A^{\mathrm{raw}}_{i,t}
    =
    \begin{cases}
        A^{\mathrm{down}}_{i,t}, & B_i>0\ \text{and}\ g_{i,t}<0,\\
        A^{\mathrm{up}}_{i,t}, & B_i<0\ \text{and}\ g_{i,t}>0,\\
        \hat A_{i,t}, & \text{otherwise}.
    \end{cases}
\end{equation}

Finally, token advantages are clipped for numerical stability:
\begin{equation}
\begin{aligned}
    \tilde{A}_{i,t} &= \mathrm{clip}\!\bigl(A^{\mathrm{raw}}_{i,t}, A_{\min}, A_{\max}\bigr),
\end{aligned}
\end{equation}
with default $(A_{\min},A_{\max})=(-1.2,1.2)$.

This token-level rule is the key difference between CAST and correctness-unaware self-distillation. A correct trajectory is not treated as uniformly positive at every token: locally teacher-negative tokens can still be suppressed. Likewise, an incorrect trajectory is not treated as uniformly negative at every token: locally teacher-positive tokens can receive bounded positive credit, while the trajectory-level verifier still determines the base sign and keeps the update grounded in outcome correctness.

\subsection{Optimization Objective}

CAST uses the DAPO/GRPO clipped importance-ratio objective with token-level advantages~\cite{ppo,grpo,dapo}:
\begin{equation}
    \label{eq:cast_objective}
    \mathcal{L}_{\mathrm{CAST}}
    =
    -\mathbb{E}_{i,t}
    \left[
    \min
    \left(
    \rho_{i,t}\tilde A_{i,t},
    \mathrm{clip}
    \left(
    \rho_{i,t},
    1-\epsilon,
    1+\epsilon_{\mathrm{high}}
    \right)
    \tilde A_{i,t}
    \right)
    \right],
\end{equation}
where
\begin{equation}
\rho_{i,t}
=
\frac{\pi_\theta(y_{i,t}\mid x,y_{i,<t})}
{\pi_{\theta_{\mathrm{old}}}(y_{i,t}\mid x,y_{i,<t})}.
\end{equation}
The policy objective remains verifier-grounded at the trajectory level: rewards determine trajectory correctness, group type, and the sign of the base advantage. Answer-free self-teacher scoring then refines token-level credit assignment through the gap $g_{i,t}$, and may assign negative advantages to teacher-negative tokens within positive-base trajectories or bounded positive advantages to teacher-positive tokens within negative-base trajectories. The shaped advantages are detached when optimizing this objective, so CAST does not introduce gradients through the teacher-scoring computation.

Implementation details for the detached advantage construction are provided in Appendix~\ref{app:implementation_details}.

\begin{table}[h]
  \centering
  \setlength{\tabcolsep}{4pt}
  \renewcommand{\arraystretch}{1.08}
  \begin{adjustbox}{max width=\textwidth}
  \begin{tabular}{@{}lcccc@{}}
    \toprule
    \textbf{Method}
    & \textbf{Token advantage}
    & \textbf{Zero-variance groups}
    & \textbf{Privileged information}
    & \textbf{Leakage risk} \\
    \midrule
    GRPO / DAPO
    & Sequence-level
    & No
    & No
    & No \\
    OPSD
    & Token-level
    & No
    & Yes
    & Yes \\
    GRPO+OPSD
    & Sequence--token
    & No
    & Yes
    & Yes \\
    RLSD / SDPO-style
    & Token-level
    & No
    & Yes
    & Yes \\
    RLRT
    & Token-level
    & No
    & Yes
    & Yes \\
    CAST
    & Token-level
    & Yes
    & No
    & No \\
    \bottomrule
  \end{tabular}
  \end{adjustbox}
  \caption{Conceptual comparison between CAST and related GRPO/self-distillation variants, under the implementations considered in this paper.}
  \label{tab:method_comparison}
\end{table}

The full training procedure is given in Appendix Algorithm~\ref{alg:cast}.

\section{Experiments}
\subsection{Training Details}

We train Qwen3-1.7B, Qwen3-4B, and Qwen3-8B~\cite{qwen3} on DAPO-Math-17K~\cite{dapo} for 300 optimizer steps under a matched rollout and optimization budget. We verify no exact problem-text overlap after normalization between DAPO-Math-17K and the in-domain evaluation benchmarks (AIME24, AIME25, AIME26, MATH-500, and HMMT25). The main comparison spans all three model sizes, while Qwen3-4B is used for mechanistic ablations and diagnostics. All runs use LoRA~\cite{lora}. The RLVR-style methods use binary verifier rewards, 8 completions per prompt, and 128 on-policy trajectories per update; the OPSD-only baseline instead optimizes the reference-solution-conditioned self-distillation objective.

All main runs fit on two NVIDIA A800 80GB GPUs and use only a few hundred optimizer steps. The comparison includes GRPO, OPSD, GRPO+OPSD, RLSD, RLRT~\cite{rlrt}, and CAST. Here, OPSD denotes the standard on-policy self-distillation objective without a verifier-grounded GRPO term. The main method comparison uses a matched 300-step budget; 600-step CAST checkpoints are reported later only as scaling evidence under longer training, not as the main matched-budget comparison. GRPO+OPSD denotes an additive baseline,
\[
\mathcal{L}_{\mathrm{GRPO+OPSD}}
=
\mathcal{L}_{\mathrm{GRPO}}
\;+\;
\alpha_k\,\mathcal{L}_{\mathrm{OPSD}},
\]
where \(\alpha_k\) is the OPSD auxiliary coefficient. This baseline does not use CAST's correctness-aware group splitting, uniform-reward base-advantage branch, answer-free self-teacher scoring, or bidirectional advantage sign reversal. Full optimizer, decoding, hardware, and method-specific hyperparameters are provided in Appendix~\ref{app:training_details}; additional training diagnostics are provided in Appendix~\ref{app:training_diagnostics}.

Figure~\ref{fig:main_training_dynamics} shows training dynamics for the main Qwen3-4B runs. These curves are not used as final evaluation metrics, but they summarize the optimization behavior of CAST: response length and entropy provide stability checks, while token-advantage magnitude reflects the strength of token-level feedback.

\begin{figure}[h]
  \centering
  \begin{minipage}[t]{0.32\textwidth}
    \centering
    \includegraphics[width=\linewidth]{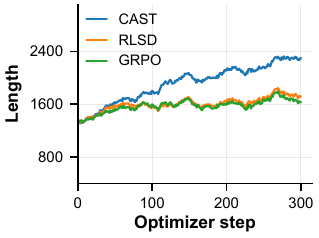}
  \end{minipage}
  \hfill
  \begin{minipage}[t]{0.32\textwidth}
    \centering
    \includegraphics[width=\linewidth]{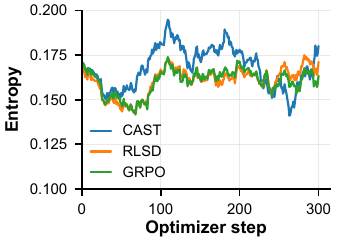}
  \end{minipage}
  \hfill
  \begin{minipage}[t]{0.32\textwidth}
    \centering
    \includegraphics[width=\linewidth]{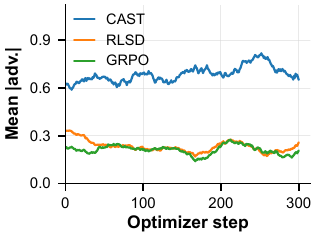}
  \end{minipage}
  \caption{Training dynamics for Qwen3-4B CAST, RLSD, and GRPO over 300 optimizer steps. Response length and entropy summarize generation length and exploration behavior during training, while token-advantage magnitude reflects the strength of token-level policy feedback.}
  \label{fig:main_training_dynamics}
\end{figure}

\subsection{Results}

For evaluation, each LoRA adapter is loaded on top of its corresponding Qwen3 base model and evaluated with vLLM in the non-thinking generation mode. Here, ``non-thinking'' means disabling the model-specific hidden thinking mode; the model is still prompted to produce visible step-by-step reasoning text. Following the multi-sample evaluation practice common in mathematical reasoning~\cite{self_consistency,minerva}, the protocol samples 16 responses per problem on AIME24\footnote{\url{https://huggingface.co/datasets/math-ai/aime24}}, AIME25\footnote{\url{https://huggingface.co/datasets/math-ai/aime25}}, AIME26\footnote{\url{https://huggingface.co/datasets/math-ai/aime26}}, MATH-500~\cite{math_dataset}\footnote{\url{https://huggingface.co/datasets/HuggingFaceH4/MATH-500}}, and HMMT25\footnote{\url{https://huggingface.co/datasets/MathArena/hmmt_feb_2025}}. Pass@16 measures best-of-16 coverage, and Avg@16 measures average per-sample accuracy. Full prompt, decoding, and verification details are provided in Appendix~\ref{app:evaluation_details}.

Table~\ref{tab:main_avg_pass16} summarizes the main results across Qwen3-1.7B, Qwen3-4B, and Qwen3-8B. The exploratory RLSD variant that also updates all-correct and all-wrong groups, denoted as RLSD (with zero-variance updates), is moved to Appendix~\ref{app:additional_explorations} rather than included as a main baseline.

\begin{table}[!htbp]
  \centering
  \setlength{\tabcolsep}{3.0pt}
  \renewcommand{\arraystretch}{1.06}
  \begin{adjustbox}{max width=0.96\textwidth}
  \begin{tabular}{@{}lcccccccccc@{}}
    \toprule
    \textbf{Method}
    & \multicolumn{2}{c}{AIME24}
    & \multicolumn{2}{c}{AIME25}
    & \multicolumn{2}{c}{AIME26}
    & \multicolumn{2}{c}{MATH-500}
    & \multicolumn{2}{c}{HMMT25} \\
    \cmidrule(lr){2-3}
    \cmidrule(lr){4-5}
    \cmidrule(lr){6-7}
    \cmidrule(lr){8-9}
    \cmidrule(lr){10-11}
    & Avg & Pass & Avg & Pass & Avg & Pass & Avg & Pass & Avg & Pass \\
    \midrule
    \multicolumn{11}{@{}l}{\textbf{Qwen3-1.7B}} \\
    Base & 12.71 & 43.3 & 9.58 & 33.3 & 9.38 & 23.3 & 73.15 & 92.4 & 6.04 & 26.7 \\
    +GRPO & 14.17 & 40.0 & 9.38 & 33.3 & 10.00 & 30.0 & 72.90 & 93.4 & 4.38 & 10.0 \\
    +OPSD & 5.21 & 13.3 & 7.50 & 26.7 & 2.92 & 13.3 & 68.91 & 91.6 & 1.88 & 10.0 \\
    +GRPO+OPSD & 13.33 & 33.3 & 11.04 & 23.3 & 7.50 & 23.3 & 72.86 & 91.6 & 5.62 & 26.7 \\
    +RLSD & 12.50 & 36.7 & 9.58 & 33.3 & 11.04 & 33.3 & 73.21 & 92.0 & 6.04 & 26.7 \\
    +RLRT & 11.67 & 36.7 & 9.79 & 23.3 & 9.38 & 26.7 & 74.00 & 92.6 & 5.62 & 16.7 \\
    +CAST & \textbf{24.58} & \textbf{56.7} & \textbf{19.79} & \textbf{43.3} & \textbf{17.29} & \textbf{36.7} & \textbf{80.80} & \textbf{95.6} & \textbf{13.12} & \textbf{30.0} \\
    \midrule
    \multicolumn{11}{@{}l}{\textbf{Qwen3-4B}} \\
    Base & 20.83 & 46.7 & 21.25 & 50.0 & 17.71 & 50.0 & 83.91 & 96.0 & 10.21 & 23.3 \\
    +GRPO & 21.04 & 43.3 & 22.50 & 50.0 & 20.00 & 53.3 & 83.86 & 95.4 & 12.08 & 30.0 \\
    +OPSD & 10.62 & 30.0 & 3.75 & 23.3 & 9.17 & 13.3 & 72.35 & 86.2 & 2.29 & 10.0 \\
    +GRPO+OPSD & 22.50 & 53.3 & 21.67 & 46.7 & 17.29 & 50.0 & 83.74 & 96.0 & 12.92 & 23.3 \\
    +RLSD & 21.67 & 50.0 & 21.88 & 50.0 & 17.71 & 40.0 & 84.08 & 96.2 & 11.88 & 23.3 \\
    +RLRT & 21.67 & 53.3 & 22.08 & 46.7 & 20.00 & 63.3 & 84.35 & 95.6 & 12.92 & 26.7 \\
    +CAST & \textbf{41.25} & \textbf{76.7} & \textbf{35.83} & \textbf{70.0} & \textbf{36.25} & \textbf{70.0} & \textbf{89.38} & \textbf{98.0} & \textbf{21.67} & \textbf{50.0} \\
    \midrule
    \multicolumn{11}{@{}l}{\textbf{Qwen3-8B}} \\
    Base & 24.79 & 50.0 & 19.79 & 43.3 & 14.58 & 43.3 & 83.94 & 96.8 & 10.83 & 26.7 \\
    +GRPO & 27.08 & 56.7 & 22.29 & 46.7 & 15.21 & 40.0 & 83.85 & 96.4 & 9.58 & 30.0 \\
    +OPSD & 15.62 & 33.3 & 16.25 & 30.0 & 13.75 & 40.0 & 76.69 & 91.0 & 7.08 & 20.0 \\
    +GRPO+OPSD & 27.29 & 56.7 & 20.83 & 43.3 & 16.67 & 46.7 & 83.78 & 96.6 & 11.25 & 30.0 \\
    +RLSD & 26.25 & 63.3 & 20.83 & 46.7 & 17.29 & 50.0 & 84.04 & 96.4 & 11.88 & 26.7 \\
    +RLRT & 26.67 & 60.0 & 23.75 & 46.7 & 17.71 & 53.3 & 84.06 & 96.2 & 13.33 & 30.0 \\
    +CAST & \textbf{42.29} & \textbf{73.3} & \textbf{32.29} & \textbf{70.0} & \textbf{32.29} & \textbf{66.7} & \textbf{88.74} & \textbf{97.6} & \textbf{18.96} & \textbf{43.3} \\
    \bottomrule
  \end{tabular}
  \end{adjustbox}
  \caption{Avg@16 and Pass@16 performance (\%) across model scales. Methods are grouped under each Qwen3 model size.}
  \label{tab:main_avg_pass16}
\end{table}

\begin{figure}[!htbp]
  \centering
  \includegraphics[width=0.96\linewidth]{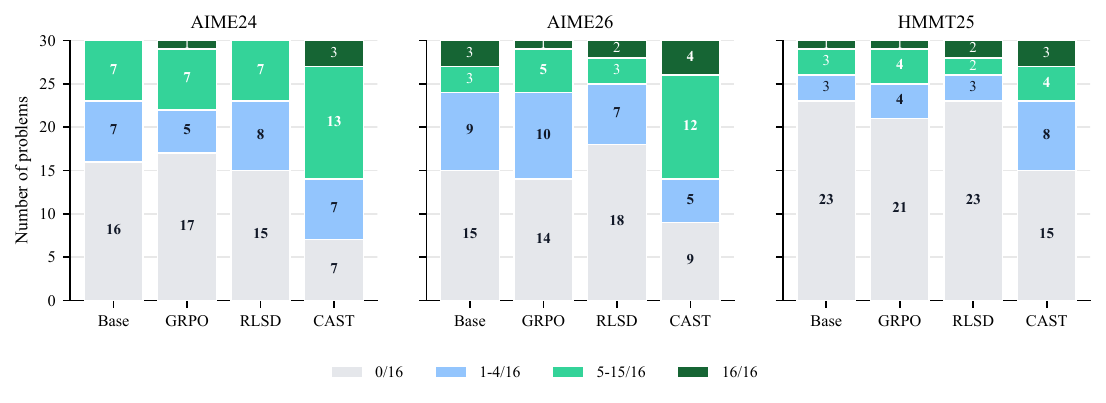}
  \caption{Per-problem distribution of correct samples across AIME24, AIME26, and HMMT25. Each bar partitions the 30 problems in each benchmark by how many of the 16 sampled responses are correct.}
  \label{fig:problem_bucket_comparison}
\end{figure}

Figure~\ref{fig:problem_bucket_comparison} examines the per-problem distribution of the 16 sampled responses on Qwen3-4B across AIME24, AIME26, and HMMT25. For each problem \(j\), let \(c_j\in\{0,\ldots,16\}\) denote the number of verified-correct samples; the figure compares the distribution of \(c_j\) across the base model, GRPO, RLSD, and CAST.

Across model scales, CAST improves Avg@16 and Pass@16 over the base model. In the Qwen3-4B setting, CAST outperforms GRPO, OPSD, GRPO+OPSD, RLSD, and RLRT on every benchmark under both metrics. The same matched method set is evaluated on Qwen3-1.7B and Qwen3-8B, where CAST gives the strongest overall Avg@16 and competitive Pass@16, with the highest Pass@16 in most comparisons. CAST also produces substantially longer completions on average; length statistics and token-budget sweeps are reported in Appendix Tables~\ref{tab:app_eval_output_length} and~\ref{tab:app_token_budget_avg_pass16}. Its gains are strongest under sufficiently large generation budgets, while under tight token budgets the advantage can shrink or disappear.

CAST shifts probability mass away from the all-failure (0/16) stratum and toward medium-to-high success strata. Relative to the base model, all-failure problems decrease from 16 to 7 on AIME24, from 15 to 9 on AIME26, and from 23 to 15 on HMMT25, while the combined 5--15/16 and 16/16 strata increase from 7 to 16, from 6 to 16, and from 4 to 7, respectively.

\section{Ablation}

\subsection{Self-Teacher Shaping Cutoff}

CAST uses answer-free self-teacher scoring. This ablation varies when the CAST-specific token shaping is removed. For cutoff step \(K\), the token-gap coefficient is kept constant before \(K\) and set to zero afterward:
\begin{equation}
\lambda_k^{\mathrm{cutoff}}
=
\begin{cases}
\lambda_0, & k<K,\\
0, & k\ge K.
\end{cases}
\end{equation}
For this cutoff ablation, the uniform-reward base branch is also gated after step \(K\):
\begin{equation}
b^{\mathrm{cutoff}}_{\mathrm{correct},k}
=
b^{\mathrm{cutoff}}_{\mathrm{wrong},k}
=
\begin{cases}
1, & k<K,\\
0, & k\ge K.
\end{cases}
\end{equation}
This schedule imposes a hard cutoff. The ablation tests whether CAST benefits from persistent correctness-aware token shaping. Once \(k\ge K\), teacher--student gaps no longer affect token-level advantages, and the uniform-reward base advantages are also disabled. Thus, after the cutoff, all-correct and all-wrong groups no longer receive CAST-specific fixed base-advantage updates. With zero-based optimizer indices $k=0,\ldots,299$ over 300 updates, the cutoff-at-300 setting ($K=300$) keeps both teacher-gap shaping and uniform-reward base-advantage updates active for every update.

\begin{table}[h]
  \centering
  \setlength{\tabcolsep}{3.0pt}
  \renewcommand{\arraystretch}{1.06}
  \begin{adjustbox}{max width=0.96\textwidth}
  \begin{tabular}{@{}llcccccccccc@{}}
    \toprule
    \textbf{Schedule} & \textbf{Step}
    & \multicolumn{2}{c}{AIME24}
    & \multicolumn{2}{c}{AIME25}
    & \multicolumn{2}{c}{AIME26}
    & \multicolumn{2}{c}{MATH-500}
    & \multicolumn{2}{c}{HMMT25} \\
    \cmidrule(lr){3-4}
    \cmidrule(lr){5-6}
    \cmidrule(lr){7-8}
    \cmidrule(lr){9-10}
    \cmidrule(lr){11-12}
    & & Avg & Pass & Avg & Pass & Avg & Pass & Avg & Pass & Avg & Pass \\
    Cutoff & 50  & 23.75 & 46.7 & 25.21 & 53.3 & 18.96 & 46.7 & 84.38 & 96.6 & 10.83 & 20.0 \\
    Cutoff & 150 & 32.92 & 70.0 & 29.79 & 60.0 & 25.42 & 53.3 & 87.50 & 97.4 & 16.88 & 33.3 \\
    Cutoff & 300 & \textbf{41.25} & \textbf{76.7} & \textbf{35.83} & \textbf{70.0} & \textbf{36.25} & \textbf{70.0} & \textbf{89.38} & \textbf{98.0} & \textbf{21.67} & \textbf{50.0} \\ 
    \bottomrule
  \end{tabular}
  \end{adjustbox}
  \caption{Effect of the self-teacher shaping cutoff on Avg@16 and Pass@16 performance of CAST.}
  \label{tab:ablation_cast_schedule_avg_pass16}
\end{table}

\subsection{Effect of Training Steps Across Model Scales}

This ablation evaluates whether CAST's improvements are sensitive to the optimizer step at which checkpoints are selected. The default CAST configuration is kept fixed, including answer-free self-teacher scoring, non-decayed token-gap shaping, uniform-reward base-advantage updates, and asymmetric token-weight clipping. For Qwen3-1.7B, Qwen3-4B, and Qwen3-8B, the original base model is compared with CAST checkpoints trained for 100, 300, and 600 optimizer steps under the same evaluation protocol. The 300-step checkpoints correspond to the default settings used in the main comparison.

\begin{table}[h]
  \centering
  \setlength{\tabcolsep}{2.6pt}
  \renewcommand{\arraystretch}{1.06}
  \begin{adjustbox}{max width=0.96\textwidth}
  \begin{tabular}{@{}llcccccccccc@{}}
    \toprule
    \textbf{Model} & \textbf{Checkpoint}
    & \multicolumn{2}{c}{AIME24}
    & \multicolumn{2}{c}{AIME25}
    & \multicolumn{2}{c}{AIME26}
    & \multicolumn{2}{c}{MATH-500}
    & \multicolumn{2}{c}{HMMT25} \\
    \cmidrule(lr){3-4}
    \cmidrule(lr){5-6}
    \cmidrule(lr){7-8}
    \cmidrule(lr){9-10}
    \cmidrule(lr){11-12}
    & & Avg & Pass & Avg & Pass & Avg & Pass & Avg & Pass & Avg & Pass \\
    \midrule
    Qwen3-1.7B & Base & 12.71 & 43.3 & 9.58 & 33.3 & 9.38 & 23.3 & 73.15 & 92.4 & 6.04 & 26.7 \\
    Qwen3-1.7B & 100 & 17.92 & 43.3 & 13.3 & 30.0 & 10.21 & 36.7 & 76.09 & 93.6 & 7.08 & 20.0 \\
    Qwen3-1.7B & 300 & 24.58 & 56.7 & 19.79 & 43.3 & 17.29 & 36.7 & 80.80 & 95.6 & 13.12 & 30.0 \\
    Qwen3-1.7B & 600 & \textbf{43.12} & \textbf{73.3} & \textbf{37.71} & \textbf{56.7} & \textbf{37.50} & \textbf{63.3} & \textbf{89.56} & \textbf{97.6} & \textbf{23.33} & \textbf{46.7} \\
    \midrule
    Qwen3-4B & Base & 20.83 & 46.7 & 21.25 & 50.0 & 17.71 & 50.0 & 83.91 & 96.0 & 10.21 & 23.3 \\
    Qwen3-4B & 100 & 28.33 & 60.0 & 24.38 & 50.0 & 21.88 & 53.3 & 85.61 & 96.8 & 13.96 & 40.0 \\
    Qwen3-4B & 300 & 41.25 & 76.7 & 35.83 & 70.0 & 36.25 & 70.0 & 89.38 & 98.0 & 21.67 & 50.0 \\
    Qwen3-4B & 600 & \textbf{56.46} & \textbf{83.3} & \textbf{52.92} & \textbf{76.7} & \textbf{50.00} & \textbf{76.7} & \textbf{93.10} & \textbf{98.2} & \textbf{27.92} & \textbf{56.7} \\
    \midrule
    Qwen3-8B & Base & 24.79 & 50.0 & 19.79 & 43.3 & 14.58 & 43.3 & 83.94 & 96.8 & 10.83 & 26.7 \\
    Qwen3-8B & 100 & 30.21 & 63.3 & 21.88 & 43.3 & 17.92 & 50.0 & 85.08 & 97.2 & 12.08 & 30.0 \\
    Qwen3-8B & 300 & 42.29 & 73.3 & 32.29 & 70.0 & 32.29 & 66.7 & 88.74 & 97.6 & 18.96 & 43.3 \\
    Qwen3-8B & 600 & \textbf{54.79} & \textbf{86.7} & \textbf{51.88} & \textbf{83.3} & \textbf{45.62} & \textbf{80.0} & \textbf{91.40} & \textbf{98.2} & \textbf{26.04} & \textbf{63.3} \\
    \bottomrule
  \end{tabular}
  \end{adjustbox}
  \caption{Effect of training duration on Avg@16 and Pass@16 performance of CAST across model scales. Base denotes the original model before CAST training; within each model size, CAST rows use the same configuration and differ only in the number of optimizer steps.}
  \label{tab:ablation_training_steps}
\end{table}

\subsection{Component Ablations}

This section isolates the main CAST design choices in the Qwen3-4B, 300-step setting. Table~\ref{tab:ablation_component_matrix} makes the ablation switches explicit, and Table~\ref{tab:ablation_components} reports all corresponding results in a single table. All variants keep the same optimizer, rollout budget, data, verifier, LoRA configuration, and evaluation protocol; each row changes only the component indicated by the matrix. The following subsubsections discuss each component so that the aggregate table can be read as a controlled decomposition rather than as a collection of unrelated variants.

\begin{table}[h]
  \centering
  \setlength{\tabcolsep}{4pt}
  \renewcommand{\arraystretch}{1.06}
  \begin{adjustbox}{max width=\textwidth}
  \begin{tabular}{@{}lcccccc@{}}
    \toprule
    \textbf{Variant}
    & \textbf{Mixed-group shaping}
    & \textbf{Zero-variance branch}
    & \textbf{Answer-free scoring}
    & \textbf{Asymmetric clip}
    & \textbf{Correct-path negative reversal}
    & \textbf{Incorrect-path positive reversal} \\
    \midrule
    CAST & \good & \good & \good & \good & \good & \good \\
    Mixed-only & \good & \reject & \good & \good & \good & \good \\
    With ground-truth answer & \good & \good & \reject & \good & \good & \good \\
    Sym. $[0.95,1.05]$ & \good & \good & \good & \reject & \good & \good \\
    Sym. $[0.8,1.2]$ & \good & \good & \good & \reject & \good & \good \\
    No sign reversal & \good & \good & \good & \good & \reject & \reject \\
    \bottomrule
  \end{tabular}
  \end{adjustbox}
  \caption{Component switches for Table~\ref{tab:ablation_components}. ``Mixed-group shaping'' denotes the standard mixed-group GRPO base advantage combined with CAST token-gap shaping. ``Zero-variance branch'' denotes bounded uniform-reward base advantages on all-correct/all-wrong groups with the same token-level shaping rule. ``Correct-path negative reversal'' applies sign reversal to teacher-negative positive-base tokens; ``Incorrect-path positive reversal'' applies bounded positive sign reversal to teacher-positive negative-base tokens.}
  \label{tab:ablation_component_matrix}
\end{table}

\begin{table}[h]
  \centering
  \setlength{\tabcolsep}{2.3pt}
  \renewcommand{\arraystretch}{1.06}
  \begin{adjustbox}{max width=\textwidth}
  \begin{tabular}{@{}llcccccccccc@{}}
    \toprule
    \textbf{Ablation} & \textbf{Variant}
    & \multicolumn{2}{c}{AIME24}
    & \multicolumn{2}{c}{AIME25}
    & \multicolumn{2}{c}{AIME26}
    & \multicolumn{2}{c}{MATH-500}
    & \multicolumn{2}{c}{HMMT25} \\
    \cmidrule(lr){3-4}
    \cmidrule(lr){5-6}
    \cmidrule(lr){7-8}
    \cmidrule(lr){9-10}
    \cmidrule(lr){11-12}
    & & Avg & Pass & Avg & Pass & Avg & Pass & Avg & Pass & Avg & Pass \\
    \midrule
    Default & CAST 
    & \textbf{41.25} & \textbf{76.7} & \textbf{35.83} & \textbf{70.0} & 36.25 & 70.0 & \textbf{89.38} & \textbf{98.0} & \textbf{21.67} & \textbf{50.0} \\
    \midrule
    Zero-variance groups & Mixed-only  & 34.38 & \textbf{76.7} & 31.67 & 60.0 & 30.21 & 60.0 & 88.25 & 97.4 & 16.88 & 33.3 \\
    \midrule
    Teacher context & With ground-truth answer & 36.25 & \textbf{76.7} & 28.54 & 66.7 & 28.75 & 66.7 & 88.29 & 97.2 & 17.08 & \textbf{50.0} \\
    \midrule
    Token clipping & Sym. $[0.8,1.2]$  & 37.92 & 70.0 & 35.21 & \textbf{70.0} & 32.29 & 66.7 & 89.03 & 97.8 & 20.21 & 46.7 \\
    Token clipping & Sym. $[0.95,1.05]$   & 38.75 & 73.3 & 33.33 & \textbf{70.0} & \textbf{36.67} & \textbf{73.3} & 89.26 & 97.8 & 20.83 & \textbf{50.0} \\
    \midrule
    Sign reversal & No sign reversal 
    & 25.83 & 56.7 
    & 24.17 & 40.0 
    & 21.04 & 53.3 
    & 85.00 & 96.4 
    & 17.50 & 40.0 \\
    \bottomrule
  \end{tabular}
  \end{adjustbox}
  \caption{Component ablations for CAST in the Qwen3-4B, 300-step setting. The first row is the default CAST configuration; subsequent rows ablate one component at a time. All rows report Avg@16 and Pass@16 under the same evaluation protocol.}
  \label{tab:ablation_components}
\end{table}

\subsubsection{Zero-Variance Group Updates}

The \emph{Mixed-only} variant keeps CAST's answer-free self-teacher gap, asymmetric clipping, and bidirectional sign reversal on mixed groups, but disables the all-correct/all-wrong uniform-reward branch. Thus, Mixed-only versus CAST tests whether zero-variance groups add useful signal beyond mixed-group token shaping.

\subsubsection{Teacher Scoring Context}

The default CAST configuration uses answer-free self-teacher scoring (Appendix~\ref{app:implementation_details}). The \emph{With ground-truth answer} ablation instead conditions the teacher on the dataset reference solution $y^\ast$ (the verified final-answer string from training metadata, identical to the verifier's ground truth) before scoring the on-policy rollout:
\[
T_{\mathrm{GT}}(x,y^\ast)
=
\texttt{\{x\}}
\;\Vert\;
\texttt{"\textbackslash n\textbackslash n[Reference solution]\textbackslash n"}
\;\Vert\;
\texttt{\{solution\}\textbackslash n\textbackslash n[Student response]\textbackslash n"},
\]
where \texttt{solution} is set to $y^\ast$, $\Vert$ denotes string concatenation, and, at token $y_{i,t}$, the teacher is evaluated on the prefix $T_{\mathrm{GT}}(x,y^\ast)\Vert y_{i,<t}$. This matches the privileged reference-solution teacher template used in OPSD-style training, while keeping CAST's correctness-aware token shaping and zero-variance branch. The ablation tests whether answer-free self-teacher scoring is necessary or whether this privileged context provides comparable token-level guidance.

\subsubsection{Token-Weight Clipping}

The two symmetric-clipping rows replace CAST's asymmetric token-weight ranges with symmetric alternatives. This ablation tests whether the proposed positive- and negative-base clipping asymmetry contributes beyond using a bounded teacher-gap weight.

\subsubsection{Bidirectional Advantage Sign Reversal}

The \emph{No sign reversal} variant keeps the uniform-reward branch, answer-free self-teacher scoring, and asymmetric clipping, but prevents local sign changes in both directions. Thus, teacher-negative positive-base tokens cannot become negative, and teacher-positive negative-base tokens cannot become positive. This ablation isolates the role of two-sided local correction beyond sign-preserving token-gap weighting.

\section{Limitations}
Several limitations remain.

\begin{itemize}
    \item \textbf{Scale.}
    The experiments evaluate Qwen3-1.7B, Qwen3-4B, and Qwen3-8B with LoRA and up to 600 optimizer steps. Larger models, longer training, full-parameter updates, and broader model families remain untested.

    \item \textbf{Diagnostic scope.}
    Mechanistic analyses are limited primarily to the 300-step Qwen3-4B setting and an OPSD diagnostic set. Broader answer-free self-teacher-gap diagnostics would further test the mechanism.

    \item \textbf{Non-thinking evaluation.}
    CAST is trained and evaluated with extended reasoning disabled. Explicit chain-of-thought generation is left to future work.

    \item \textbf{Efficiency and sensitivity.}
    CAST uses validation-tuned clipping and token-gap hyperparameters, and it often produces longer outputs. This may improve exploration but also increases inference cost; length- or compute-normalized evaluation would be useful.
\end{itemize}

\section{Conclusion}

This work introduced CAST (Non-Privileged Clipped Asymmetric Self-Teaching with Advantage Flipping for GRPO). CAST combines verifier-grounded trajectory correctness with answer-free self-teacher scoring, allowing token-level advantages to depend on both rollout correctness and teacher--student gaps. This design can suppress locally teacher-negative tokens in correct rollouts and provide bounded positive credit to locally teacher-positive tokens in incorrect rollouts, requires no reference reasoning traces, and does not expose the ground-truth answer or reference solution to the self-teacher.

Under a matched rollout and optimization budget, CAST achieves the strongest overall Avg@16 in the comparisons and improves Pass@16 over GRPO, OPSD, GRPO+OPSD, RLSD, and RLRT on mathematical reasoning benchmarks. The results suggest that correctness-aware token shaping can make self-distilled RLVR more reliable, especially when zero-variance groups are explicitly used rather than discarded.

\newpage

\nocite{process_outcome_feedback,self_verification,wizardmath}

\printbibliography[title={References}]

\newpage
\section{Appendix}

\begin{algorithm}[h]
    \begin{algorithmic}[1]
    \Require Online policy $\pi_\theta$, rollout/reference policy $\pi_{\theta_{\mathrm{old}}}$, stop-gradient self-teacher $\pi_\phi$, verifier $\mathcal{V}$, prompts $\mathcal{D}$, group size $G$
    \Require Shaping coefficient $\lambda$, base advantages $b_{\mathrm{correct}},b_{\mathrm{wrong}}$
    \Require Weight clips $(\ell_+,u_+)$ and $(\ell_-,u_-)$, token-advantage clip range $(A_{\min},A_{\max})$

    \For{each training step $k$}
        \State Set rollout/reference policy $\pi_{\theta_{\mathrm{old}}}\leftarrow \mathrm{sg}(\pi_\theta)$
        \State Sample prompts $\{x_b\}\sim\mathcal{D}$

        \For{each prompt $x_b$}
            \State Sample rollouts $\{y_i\}_{i=1}^{G}\sim\pi_{\theta_{\mathrm{old}}}(\cdot\mid x_b)$
            \State Compute verifier rewards $r_i=\mathcal{V}(x_b,y_i)$
            \State Classify group type $\tau_b\in\{\textsc{Mixed},\textsc{AllCorrect},\textsc{AllWrong}\}$

            \If{$\tau_b=\textsc{Mixed}$}
                \State Compute GRPO advantages $A_i$
            \EndIf

            \For{each trajectory $y_i$}
                \State Set base advantage
                \[
                B_i \gets
                \begin{cases}
                A_i, & \tau_b=\textsc{Mixed},\\
                +b_{\mathrm{correct}}, & \tau_b=\textsc{AllCorrect},\\
                -b_{\mathrm{wrong}}, & \tau_b=\textsc{AllWrong}.
                \end{cases}
                \]

                \For{each valid token $y_{i,t}$}
                    \State Compute answer-free self-teacher gap
                    \[
                    g_{i,t}\gets \mathrm{sg}\Big(
                    \log\pi_\phi(y_{i,t}\mid x_b,y_{i,<t})
                    -
                    \log\pi_{\theta_{\mathrm{old}}}(y_{i,t}\mid x_b,y_{i,<t})
                    \Big)
                    \]
                    \State $z_{i,t}\gets \exp(\operatorname{sign}(B_i)g_{i,t})$

                    \If{$B_i>0$}
                        \State $w_{i,t}\gets \mathrm{clip}(z_{i,t},\ell_+,u_+)$
                    \Else
                        \State $w_{i,t}\gets \mathrm{clip}(z_{i,t},\ell_-,u_-)$
                    \EndIf

                    \If{$B_i>0$ and $g_{i,t}<0$}
                        \State $w^{-}_{i,t}\gets \mathrm{clip}(\exp(-g_{i,t}),\max(1,\ell_-),u_-)$
                        \State \mbox{$A^{\mathrm{raw}}_{i,t}\gets -\lvert B_i\rvert\bigl(1+\lambda(w^{-}_{i,t}-1)\bigr)$}
                    \ElsIf{$B_i<0$ and $g_{i,t}>0$}
                        \State $w^{+}_{i,t}\gets \mathrm{clip}(\exp(g_{i,t}),\max(1,\ell_+),u_+)$
                        \State \mbox{$A^{\mathrm{raw}}_{i,t}\gets \lvert B_i\rvert\bigl(1+\lambda(w^{+}_{i,t}-1)\bigr)$}
                    \Else
                        \State \mbox{$A^{\mathrm{raw}}_{i,t}\gets B_i\left(1+\lambda(w_{i,t}-1)\right)$}
                    \EndIf

                    \State \mbox{$\tilde{A}_{i,t}\gets \mathrm{clip}\!\bigl(A^{\mathrm{raw}}_{i,t}, A_{\min}, A_{\max}\bigr)$}
                \EndFor
            \EndFor
        \EndFor

        \State Update $\pi_\theta$ using the clipped DAPO/GRPO objective with $\tilde A_{i,t}$
        \State Periodically refresh the stop-gradient self-teacher $\pi_\phi\leftarrow\pi_\theta$
    \EndFor
    \end{algorithmic}
    \caption{CAST: Non-Privileged Clipped Asymmetric Self-Teaching with Advantage Flipping for GRPO}
    \label{alg:cast}
\end{algorithm}

\subsection{Related Work Details}
\label{app:related_work_details}

\paragraph{DAPO-style GRPO variants.}
Recent GRPO variants improve training stability for long-horizon reasoning. DAPO~\cite{dapo}, for example, introduces asymmetric clipping, dynamic sampling, token-level policy-gradient loss, and overlong reward shaping. The experiments adopt its asymmetric clipping rule,
\begin{equation}
\mathrm{clip}(\rho,1-\epsilon,1+\epsilon_{\mathrm{high}}),
\qquad
\epsilon_{\mathrm{high}}>\epsilon ,
\end{equation}
which gives positive-advantage tokens a larger upper clipping range. DAPO handles zero-variance groups mainly through dynamic sampling, whereas CAST gives such groups bounded sign-constrained base advantages.

\paragraph{Zero-variance prompt handling in GRPO.}
All-correct and all-wrong groups are commonly handled before the policy update rather than inside the advantage estimator. A standard preprocessing strategy is to discard or resample groups whose sampled rewards have zero variance, because their group-normalized GRPO advantages are zero. DAPO's dynamic sampling follows this philosophy: it seeks to build updates from prompts whose sampled completions contain both correct and incorrect responses, thereby concentrating computation on prompts that provide relative ranking signal~\cite{dapo}. This is effective for gradient utilization, but it also changes the on-policy prompt distribution toward boundary-difficulty examples and leaves no direct learning signal from prompts that are currently too easy or too hard.

Recent work argues that zero-variance prompts should not necessarily be discarded. RL-ZVP/No Prompt Left Behind keeps all-correct and all-wrong prompts and constructs entropy-guided advantages so that easy prompts can still consolidate confident behavior and hard prompts can still receive exploratory pressure~\cite{zero_variance_prompts}. ERPO~\cite{erpo} similarly focuses on residual prompts whose sampled outcomes lack informative reward contrast, using an entropy-aware policy objective and adaptive sampling to reactivate these cases. Prompt Replay addresses a related sampling problem by maintaining an online prompt buffer and replaying prompts that are likely to produce informative, non-saturated rollouts~\cite{prompt_replay}. These methods operate mainly at the prompt-selection or auxiliary-shaping level. CAST is complementary: it keeps zero-variance groups in the same GRPO-style update, assigns only bounded verifier-signed base advantages, and uses the answer-free self-teacher gap to decide which tokens within those all-correct or all-wrong trajectories should be amplified or suppressed.

\paragraph{OPD and OPSD.}
On-Policy Distillation (OPD) uses an external teacher to provide token-level supervision on student-sampled trajectories, reducing off-policy mismatch between teacher-generated data and student-visited states. On-Policy Self-Distillation (OPSD) removes the external teacher by using a stop-gradient self-teacher, often with privileged contexts such as the ground-truth answer or verified solution traces~\cite{opsd}. Although OPSD supplies dense token information, its direction is teacher-driven rather than verifier-grounded. Prior work suggests that OPSD may be more reliable for compacting successful reasoning than repairing failed trajectories~\cite{opsd_compresses}, motivating correctness-aware use of self-teacher signals.

\paragraph{Reasoning self-improvement and self-training.}
CAST is also related to methods that improve reasoning from model-generated traces. Chain-of-thought prompting and self-consistency show that intermediate reasoning paths and multiple sampled solutions can substantially improve mathematical and symbolic reasoning~\cite{cot,self_consistency}. STaR bootstraps reasoning by fine-tuning on self-generated rationales that lead to correct answers~\cite{star}, while ReST generates policy samples and reuses filtered data for reinforced self-training~\cite{rest_lm}. CAST differs from these offline or prompt-time self-improvement approaches by shaping on-policy RLVR updates directly, using verifier correctness to decide the trajectory-level direction and an answer-free self-teacher gap to refine token-level credit assignment.

\paragraph{Self-distilled RLVR.}
Self-distilled RLVR methods such as RLSD, SDPO-style training, and related hidden-distillation variants seek to combine two complementary signals: the verifier reward decides whether a rollout should be encouraged or discouraged, while the teacher-student gap provides dense token-level structure inside the rollout~\cite{rlsd,sdpo,hdpo}. This design is attractive because it keeps the update anchored to outcome correctness while avoiding the extreme sparsity of sequence-level rewards. It also explains why RLSD-type baselines are stronger than OPSD-only training in these experiments: they do not merely imitate the teacher, but use the verifier to set the rollout-level direction.

Most RLSD-style objectives, however, use the teacher--student gap primarily as a magnitude modulator rather than as a source of signed token-level feedback. A typical formulation maps the gap to a bounded weight,
\begin{equation}
w_{i,t}
=
\mathrm{clip}
\left(
\exp(\mathrm{sign}(A_i)g_{i,t}),
1-\delta,
1+\delta
\right),
\end{equation}
and sets
\begin{equation}
\tilde{A}_{i,t}
=
A_i\left((1-\lambda)+\lambda w_{i,t}\right).
\end{equation}
This preserves the GRPO update sign through $A_i$ while using distillation for dense token shaping. Consequently, every token in a positive-advantage trajectory remains positive and every token in a negative-advantage trajectory remains negative; the teacher gap changes the strength, but not the local direction, of the update. This is a conservative and stable choice, but it is limited when a correct trajectory contains locally undesirable tokens or when an incorrect trajectory contains locally useful partial reasoning.

Another limitation is data utilization. Since the base sign is inherited from the group-relative advantage, all-correct and all-wrong groups become weak or receive zero group-relative advantage when $A_i=0$. These groups are common in RLVR training: easy prompts may become all-correct, while hard prompts may remain all-wrong. Discarding or ignoring them wastes verifier-labeled trajectories that could still teach consolidation or failure avoidance. Finally, some self-distillation setups score the teacher with the ground-truth answer or a reference solution in the context, creating a privileged token-scoring signal that is stronger than what the policy would have at inference time.

CAST keeps the useful verifier anchoring of RLSD-style training but changes all three design choices. First, it interprets teacher gaps through trajectory correctness and allows two-sided local sign changes: teacher-negative tokens inside correct trajectories can receive negative token-level advantages, and teacher-positive tokens inside incorrect trajectories can receive bounded positive token-level advantages. Second, it gives all-correct and all-wrong groups bounded sign-constrained base advantages, so zero-variance data remains active. Third, it uses answer-free self-teacher scoring and reserves ground-truth answers and reference solutions for verification only.

\paragraph{RLRT and advantage sign reversal.}
RLRT provides a different interpretation of the teacher--student gap~\cite{rlrt}. Instead of treating teacher-negative tokens as uniformly undesirable, it reverses the teacher signal on verified-correct trajectories so that tokens where the student departs from the teacher can be strengthened. This is motivated by the observation that a smaller or lagged teacher may suppress alternative valid reasoning paths that the student has discovered. RLRT is therefore closely related to the advantage sign-reversal motivation in CAST: both methods reject pure teacher imitation on successful rollouts. CAST differs in three ways. It uses answer-free self-teacher scoring, applies the gap asymmetrically according to both trajectory correctness and gap sign with bidirectional local sign reversal, and explicitly uses uniform-reward all-correct and all-wrong groups through bounded base advantages.

\subsection{Implementation Details for CAST Advantages}
\label{app:implementation_details}

CAST constructs token-level advantages in two separated stages. First, after rollouts and verification, teacher and student log-probabilities are computed on the sampled tokens to form a detached gap
\begin{equation}
\bar g_{i,t}
=
\mathrm{sg}\left[
\log \pi_{\phi}(y_{i,t}\mid x,y_{i,<t})
-
\log \pi_{\theta_{\mathrm{old}}}(y_{i,t}\mid x,y_{i,<t})
\right],
\end{equation}
where \(\theta_{\mathrm{old}}\) denotes the student parameters used for rollout/log-probability collection before the policy-gradient update. This detached \(\bar g_{i,t}\), together with the detached verifier reward, is used to build \(\tilde A_{i,t}\). Second, the online policy is updated only through the importance ratio \(\rho_{i,t}\) in the DAPO/GRPO objective. Thus, gradients are not propagated through the teacher logits, the old-policy student logits used in the gap, the exponentiation, or the clipping operations that define the shaped advantages.

For mixed groups, \(B_i\) is the usual group-normalized GRPO advantage. For \textsc{AllCorrect} and \textsc{AllWrong} groups, \(B_i\) is set to \(+b_{\mathrm{correct}}\) and \(-b_{\mathrm{wrong}}\), respectively. These fixed base-advantage values determine only the sign and bounded scale of the token-level update; they do not create a separate reward model and do not expose the answer to the self-teacher. Truncated completions are masked out before advantage construction, so uniform-reward base-advantage updates are applied only to valid generated tokens.

\subsection{OPSD Token-Shift Diagnostic Details}
\label{app:opsd_motivation_diagnostic}

\paragraph{Diagnostic setup.}
The token-shift study samples 256 prompts from DAPO-Math-17K with 8 on-policy completions per prompt (2{,}048 trajectories total), using the base Qwen3-4B model under the same non-thinking generation settings as the main experiments. Verifier labels split the rollouts into 926 correct and 1{,}122 incorrect trajectories. For gap analysis only, the teacher is scored with an OPSD-style privileged context that includes the reference solution; this isolates the structure of teacher--student gaps and is distinct from the answer-free self-teacher scoring used by CAST at training time.

\paragraph{Strata.}
For each generated token, define
\[
\Delta_t
=
\log p_{\mathrm{teacher}}(y_t)
-
\log p_{\mathrm{student}}(y_t).
\]
Teacher-positive tokens have \(\Delta_t>0\) and teacher-negative tokens have \(\Delta_t<0\). Combining trajectory correctness with gap sign yields four strata:
\[
\mathcal{B} \in
\{
\mathcal{B}_{\mathrm{corr}}^{+},
\mathcal{B}_{\mathrm{corr}}^{-},
\mathcal{B}_{\mathrm{inc}}^{+},
\mathcal{B}_{\mathrm{inc}}^{-}
\}.
\]

\paragraph{Gap-weighted Top-$K$ stability.}
Rather than comparing only token counts, the analysis studies the stability of token types that dominate gap mass in each stratum. For vocabulary type \(v\), aggregate gap weight is
\[
M_{\mathcal{B},v}
=
\sum_{t\in \mathcal{B},\; y_t=v}
|\Delta_t|.
\]
Tokens are ranked by \(M_{\mathcal{B},v}\). For token \(v\), define
\[
\mu_{\mathcal{B},v}
=
\mathbb{E}\left[|\Delta_t| \mid t\in \mathcal{B},\; y_t=v\right],
\qquad
\sigma^2_{\mathcal{B},v}
=
\mathrm{Var}\left[|\Delta_t| \mid t\in \mathcal{B},\; y_t=v\right].
\]
The gap-weighted raw variance is
\[
\mathrm{Var@K}(\mathcal{B})
=
\frac{
\sum_{v\in \mathrm{TopK}(\mathcal{B})}
M_{\mathcal{B},v}\sigma^2_{\mathcal{B},v}
}{
\sum_{v\in \mathrm{TopK}(\mathcal{B})}
M_{\mathcal{B},v}
},
\]
and the gap-weighted coefficient of variation is
\[
\mathrm{CV@K}(\mathcal{B})
=
\frac{
\sum_{v\in \mathrm{TopK}(\mathcal{B})}
M_{\mathcal{B},v}\sigma_{\mathcal{B},v}/\mu_{\mathcal{B},v}
}{
\sum_{v\in \mathrm{TopK}(\mathcal{B})}
M_{\mathcal{B},v}
}.
\]
Lower \(\mathrm{CV@K}\) indicates that the dominant token types in a stratum receive gap mass more repeatably. Raw variance is scale-sensitive: a token with consistently large \(|\Delta_t|\) can have high variance even when its relative variation is low.

\paragraph{Findings and token interpretations.}
Table~\ref{tab:opsd_topk_stability} in the main text reports \(K{=}10\) statistics. Two patterns motivate CAST. First, teacher-negative signals have much larger absolute magnitude than teacher-positive signals: the top-10 mean \(|\Delta|\) is \(9.782\) in \(\mathcal{B}_{\mathrm{corr}}^{-}\) versus \(0.433\) in \(\mathcal{B}_{\mathrm{corr}}^{+}\), and \(8.306\) in \(\mathcal{B}_{\mathrm{inc}}^{-}\) versus \(0.509\) in \(\mathcal{B}_{\mathrm{inc}}^{+}\). This scale difference explains why \(\mathcal{B}_{\mathrm{corr}}^{-}\) and \(\mathcal{B}_{\mathrm{inc}}^{-}\) also have larger raw Var@10.

Second, after gap-weighted normalization, teacher-negative strata are more stable. \(\mathcal{B}_{\mathrm{corr}}^{-}\) has the lowest CV@10 (\(1.436\)), and \(\mathcal{B}_{\mathrm{inc}}^{-}\) remains much lower than \(\mathcal{B}_{\mathrm{inc}}^{+}\) (\(1.896\) versus \(3.841\)). The dominant teacher-negative tokens are high-magnitude but relatively stable types such as \texttt{<|im\_end|>}, \texttt{Answer}, whitespace, and repeated format markers. In contrast, \(\mathcal{B}_{\mathrm{inc}}^{+}\) is dominated by locally plausible symbolic tokens such as \texttt{+}, \texttt{\textbackslash}, \texttt{\$}, whitespace, parentheses, and digits, indicating that teacher-positive mass in incorrect trajectories is comparatively diffuse and less suitable as uniform positive supervision.

\subsection{Training Details}
\label{app:training_details}
Unless otherwise specified, experiments use DAPO-Math-17K for 300 optimizer steps on Qwen3-1.7B, Qwen3-4B, and Qwen3-8B. DAPO-Math-17K is used only for training; we verify no exact problem-text overlap after normalization with the evaluation benchmarks (AIME24, AIME25, AIME26, MATH-500, and HMMT25), so reported results are not affected by exact-match train--test contamination. The main method comparison applies the same protocol and matched baseline set across all three model sizes, while mechanistic ablations and diagnostics use Qwen3-4B. The training-step ablation additionally evaluates CAST checkpoints at 100 and 600 steps for each model size. LoRA with rank 64 and alpha 128 is used while keeping the backbone frozen. LoRA adapters are applied to \texttt{q\_proj}, \texttt{k\_proj}, \texttt{v\_proj}, \texttt{o\_proj}, \texttt{gate\_proj}, \texttt{up\_proj}, and \texttt{down\_proj}. Each run uses two NVIDIA A800 80GB GPUs: one for vLLM rollouts and one for policy optimization. For each prompt, 8 on-policy completions are sampled. With prompt batch size 2 and gradient accumulation 8, each optimizer update contains 16 prompt groups and 128 trajectories. This resource-constrained configuration uses a rollout micro-batch of 2 prompts, and the default run stops after 300 optimizer steps rather than relying on long-horizon or large-cluster post-training.

Training uses bfloat16 precision, scaled dot-product attention (SDPA), gradient checkpointing, maximum prompt length 1024, and maximum completion length 3072. Unless otherwise specified, generation uses temperature 0.7, top-$p$ 0.95, top-$k$ 20, min-$p$ 0.0, repetition penalty 1.0, and presence penalty 0.2. Completions that reach the maximum completion length without termination are masked from the policy loss, so uniform-reward base-advantage updates apply only to valid, non-truncated completion tokens. The optimizer is AdamW with zero weight decay. The learning rate is $1\times10^{-6}$ with 5\% warmup and polynomial decay to $1\times10^{-7}$, and the maximum gradient norm is 1.0. Rewards are binary verifier correctness rewards; no additional format or repetition rewards are used. Following DAPO, all GRPO-style methods use asymmetric clipping with $\epsilon=0.2$ and $\epsilon_{\mathrm{high}}=0.28$, and set the Kullback-Leibler (KL) coefficient to 0.

For self-distillation methods, the stop-gradient teacher is refreshed every 10 optimizer steps unless a baseline uses a fixed teacher. OPSD uses the reference-solution-conditioned OPSD objective with coefficient 1.0 and no verifier-grounded GRPO term. GRPO+OPSD adds the standard OPSD token auxiliary to the GRPO objective, using an auxiliary coefficient initialized to 0.2 and linearly decayed over 50 steps; it does not use CAST's correctness-aware group handling or advantage sign reversal. RLSD uses verifier rewards to set rollout-level directions and teacher--student gaps to modulate token-level magnitudes, with token-level weights clipped to $[0.8,1.2]$. RLRT follows the reversed-teacher rule on verified-correct trajectories~\cite{rlrt}: it forms \(d_{i,t}=\log\pi_{\theta_{\mathrm{old}}}(y_{i,t}\mid x,y_{i,<t})-\log\pi_{\phi}(y_{i,t}\mid c_i,y_{i,<t})\), clips \(w_{i,t}=\exp(\mathrm{sign}(A_i)d_{i,t})\) to \([1-\epsilon_w,1+\epsilon_w]\) with \(\epsilon_w=1.0\), and sets \(\tilde A_{i,t}=A_i[(1-\lambda_{\mathrm{RLRT}})+\lambda_{\mathrm{RLRT}}w_{i,t}]\) with \(\lambda_{\mathrm{RLRT}}=0.5\) and no decay. This reweighting is applied only to verified-correct rollouts; incorrect rollouts retain the standard GRPO sequence advantage. The RLRT teacher context uses the problem plus a successful rollout sampled for the same prompt group when available. CAST uses a token-gap coefficient of 1.0 with no decay and answer-free self-teacher scoring:
\[
T_{\mathrm{AF}}(x)
=
\texttt{\{x\}}
\;\Vert\;
\texttt{"\textbackslash n\textbackslash n[Student response]\textbackslash n"},
\]
so the teacher context excludes the ground-truth answer and reference solution. The \emph{With ground-truth answer} ablation in Table~\ref{tab:ablation_components} instead uses
\[
T_{\mathrm{GT}}(x,y^\ast)
=
\texttt{\{x\}}
\;\Vert\;
\texttt{"\textbackslash n\textbackslash n[Reference solution]\textbackslash n"}
\;\Vert\;
\texttt{\{solution\}\textbackslash n\textbackslash n[Student response]\textbackslash n"},
\]
with \texttt{solution} set to $y^\ast$ from \texttt{reward\_model.ground\_truth} on DAPO-Math-17K. In both settings, the verifier alone uses the ground truth to compute binary correctness rewards.

GRPO uses only verifier rewards and group-relative advantages. OPSD uses token-level self-distillation on on-policy rollouts without verifier rewards. GRPO+OPSD combines the GRPO policy-gradient loss with standard OPSD token distillation, but its OPSD signal remains correctness-unaware and does not reactivate zero-variance groups. RLSD combines GRPO with OPSD-style token weighting by using verifier rewards to determine rollout-level update directions and teacher--student gaps to modulate token-level magnitudes. RLRT further emphasizes exploration by treating teacher-negative tokens in correct rollouts as potentially useful departures from the teacher. CAST uses the same GRPO training loop, but interprets teacher--student gaps according to trajectory correctness. In the main CAST configuration, token shaping is applied to mixed, all-correct, and all-wrong groups; zero-variance groups receive bounded sign-constrained base advantages instead of being dropped. CAST uses asymmetric token-weight clipping with $(\ell_+,u_+)=(0.8,1.05)$ for positive-base trajectories and $(\ell_-,u_-)=(0.95,1.2)$ for negative-base trajectories, applies negative sign reversal to teacher-negative positive-base tokens and bounded positive sign reversal to teacher-positive negative-base tokens under the smaller positive-base cap, and clips final token-level advantages to $[-1.2,1.2]$.

In terms of training cost, CAST requires about $1.25\times$ the wall-clock time of GRPO for a full 300-step run. A full CAST run takes about 5 hours, compared with about 4 hours for GRPO, 4 hours and 40 minutes for RLSD, and 4 hours and 30 minutes for OPSD. The extra cost is modest and mainly comes from computing self-teacher logits for token-gap shaping and applying correctness-aware token-level advantages. See Table~\ref{tab:app_training_hparams}.

\begin{table}[h]
  \centering
  \begin{tabular}{ll}
    \toprule
    \textbf{Configuration} & \textbf{Value} \\
    \midrule
    Model sizes & Qwen3-1.7B, Qwen3-4B, and Qwen3-8B \\
    Ablation/diagnostic setting & Qwen3-4B \\
    Training data & DAPO-Math-17K \\
    Train--test overlap & No exact overlap after normalized problem-text matching \\
    Optimizer steps & 300 \\
    Optimizer & AdamW \\
    Weight decay & 0 \\
    Hardware & 2 NVIDIA A800 80GB GPUs \\
    Rollout backend & vLLM \\
    Trainable parameters & LoRA adapters only \\
    LoRA rank / alpha & 64 / 128 \\
    LoRA target modules & q, k, v, o, gate, up, down projections \\
    Precision & bfloat16 \\
    Attention implementation & SDPA \\
    Gradient checkpointing & Enabled \\
    Completions per prompt & 8 \\
    Prompt batch size & 2 \\
    Gradient accumulation & 8 \\
    Trajectories per update & 128 \\
    Max prompt / completion length & 1024 / 3072 \\
    Truncated completions & Masked from policy loss \\
    Temperature & 0.7 \\
    Top-$p$ / Top-$k$ / Min-$p$ & 0.95 / 20 / 0.0 \\
    Repetition / presence penalty & 1.0 / 0.2 \\
    Learning rate & $1\times10^{-6}$ \\
    LR schedule & 5\% warmup, polynomial decay to $1\times10^{-7}$ \\
    Max gradient norm & 1.0 \\
    RLVR reward & Binary verifier correctness \\
    KL coefficient & 0 \\
    GRPO clipping & $\epsilon=0.2$, $\epsilon_{\mathrm{high}}=0.28$ \\
    OPSD-only objective & Reference-solution-conditioned token distillation \\
    GRPO+OPSD auxiliary & OPSD coefficient 0.2, linear decay over 50 steps \\
    RLRT coefficient / decay & 0.5 / none \\
    RLRT token-weight clip & $\epsilon_w=1.0$ \\
    RLRT teacher context & Successful rollout from the same prompt group \\
    CAST teacher update interval & 10 steps \\
    CAST teacher scoring & Answer-free self-teacher scoring \\
    CAST token-gap coefficient & 1.0, no decay \\
    CAST group feedback & Mixed, all-correct, and all-wrong groups \\
    CAST positive clipping $(\ell_+,u_+)$ & $(0.8,1.05)$ \\
    CAST negative clipping $(\ell_-,u_-)$ & $(0.95,1.2)$ \\
    CAST sign reversal & Correct-path negative reversal; incorrect-path positive reversal \\
    CAST token-advantage clipping & $[-1.2,1.2]$ \\
    \bottomrule
  \end{tabular}
  \caption{Training hyperparameters used in the experiments.}
  \label{tab:app_training_hparams}
\end{table}

\subsection{Training Diagnostics}
\label{app:training_diagnostics}

Because CAST explicitly changes how uniform-reward groups and local advantage sign reversal are handled, training diagnostics from the main Qwen3-4B CAST run are also summarized. Values in Table~\ref{tab:app_training_diagnostics} are means over 300 optimizer steps. The run contains a nontrivial fraction of zero-variance prompt groups: all-correct groups account for 20.2\% of prompt groups and all-wrong groups for 31.3\%, while mixed groups account for 48.5\%. Thus, the uniform-reward branch is not an isolated edge case under this training setup.

\begin{table}[H]
  \centering
  \begin{tabular}{lc}
    \toprule
    \textbf{Quantity} & \textbf{Mean over 300 steps} \\
    \midrule
    All-correct prompt groups & 20.2\% \\
    All-wrong prompt groups & 31.3\% \\
    Mixed prompt groups & 48.5\% \\
    Sign-flipped valid tokens & 26.4\% \\
    Incorrect-path positive-flipped valid tokens & 7.0\% \\
    Truncated completion ratio & 26.8\% \\
    Mean valid completion length & 1090.6 tokens \\
    \bottomrule
  \end{tabular}
  \caption{Training diagnostics for the main Qwen3-4B CAST run. Group fractions are measured at the prompt-group level; token statistics are computed over valid completion tokens after truncation masking.}
  \label{tab:app_training_diagnostics}
\end{table}

\begin{figure*}[h]
  \centering
  \begin{subfigure}[t]{0.24\textwidth}
    \centering
    \includegraphics[width=\linewidth]{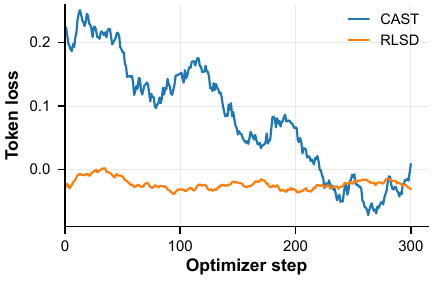}
    \caption{Token loss}
    \label{fig:app_diag_token_loss}
  \end{subfigure}
  \hfill
  \begin{subfigure}[t]{0.24\textwidth}
    \centering
    \includegraphics[width=\linewidth]{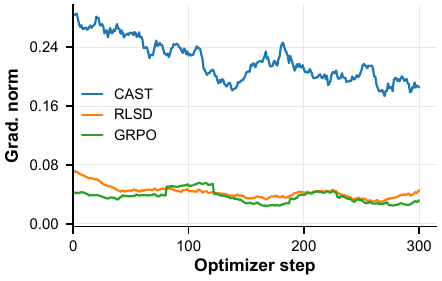}
    \caption{Gradient norm}
    \label{fig:app_diag_grad_norm}
  \end{subfigure}
  \hfill
  \begin{subfigure}[t]{0.24\textwidth}
    \centering
    \includegraphics[width=\linewidth]{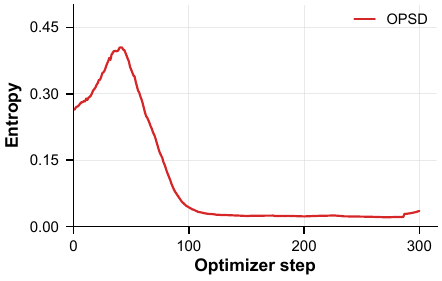}
    \caption{OPSD entropy}
    \label{fig:app_diag_opsd_entropy}
  \end{subfigure}
  \hfill
  \begin{subfigure}[t]{0.24\textwidth}
    \centering
    \includegraphics[width=\linewidth]{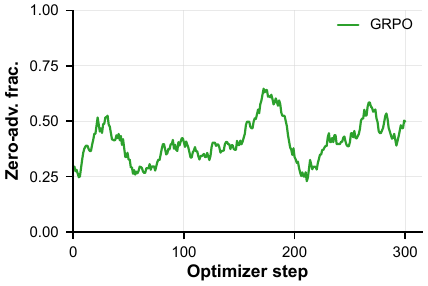}
    \caption{GRPO zero advantage}
    \label{fig:app_diag_grpo_zero_adv}
  \end{subfigure}

  \caption{Training and failure diagnostics for Qwen3-4B methods over 300 optimizer steps.}
  \label{fig:app_training_failure_diagnostics}
\end{figure*}

Figure~\ref{fig:app_training_failure_diagnostics} summarizes both optimization and failure diagnostics for Qwen3-4B methods over 300 optimizer steps. Figures~\ref{fig:app_diag_token_loss} and~\ref{fig:app_diag_grad_norm} show the token-level PPO loss term and gradient norm for CAST and RLSD, verifying that CAST's larger-magnitude token feedback remains numerically controlled. The on-policy training reward curve is omitted because it is highly batch-dependent under the small prompt budget, and benchmark performance is evaluated separately in the main result tables. Figures~\ref{fig:app_diag_opsd_entropy} and~\ref{fig:app_diag_grpo_zero_adv} show two baseline-specific failure patterns: OPSD exhibits a pronounced late-stage entropy decrease under correctness-unaware self-distillation, while GRPO often assigns zero group-relative advantage to many rollout tokens when sampled prompt groups are saturated, reducing mixed-outcome contrast and motivating CAST's bounded zero-variance branch.

\subsection{Evaluation Details}
\label{app:evaluation_details}

All methods are evaluated under a common vLLM-based protocol. The base model is evaluated directly; trained methods use the LoRA checkpoint from optimizer step 300 unless an ablation explicitly studies a different checkpoint step. Evaluation is conducted on held-out benchmarks---AIME24, AIME25, AIME26, MATH-500, and HMMT25---for which we found no exact normalized problem-text overlap with DAPO-Math-17K, reducing exact-match train--test contamination risk. For each problem, 16 completions are sampled, and Pass@$k$ is reported with $k\in\{1,4,8,16\}$. Answers are extracted from the final \texttt{\textbackslash boxed\{\}} expression and verified for mathematical equivalence with \texttt{math\_verify} when available. See Table~\ref{tab:app_eval_hparams}.

All evaluation datasets are free-form mathematical reasoning benchmarks, so a single boxed-answer prompt is used:
\[
\texttt{\{problem\}}\texttt{\textbackslash n\textbackslash nPlease reason step by step, and put your final answer within \textbackslash boxed\{\}.}
\]

\begin{table}[h]
  \centering
  \begin{tabular}{ll}
    \toprule
    \textbf{Configuration} & \textbf{Value} \\
    \midrule
    Model sizes & Qwen3-1.7B, Qwen3-4B, and Qwen3-8B \\
    Ablation/diagnostic setting & Qwen3-4B \\
    Evaluated checkpoint & LoRA checkpoint at step 300 unless otherwise specified \\
    Evaluation datasets & AIME24, AIME25, AIME26, MATH-500, HMMT25 \\
    Train--test overlap & No exact overlap after normalized problem-text matching \\
    Evaluation backend & vLLM \\
    Hardware & 2 $\times$ NVIDIA A800 80GB GPUs \\
    Tensor parallel size & 2 \\
    Precision & bfloat16 \\
    Attention backend & FlashAttention \\
    Max model length & 40960 \\
    Max new tokens & 32768 \\
    Hidden thinking mode & Disabled; visible step-by-step reasoning prompted \\
    Tokenizer / chat template & Base-model tokenizer forced \\
    Samples per problem & 16 \\
    Reported metrics & Pass@1, Pass@4, Pass@8, Pass@16 \\
    Temperature & 0.7 \\
    Top-$p$ / Top-$k$ / Min-$p$ & 0.8 / 20 / 0.0 \\
    Presence penalty & 0.0 \\
    Seed & 42 \\
    GPU memory utilization & 0.9 \\
    Generation batch size & 16 for AIME/HMMT; 64 for MATH-500 \\
    Number of evaluated examples & All examples in each benchmark \\
    Answer format & Final answer in \texttt{\textbackslash boxed\{\}} \\
    Answer verification & \texttt{math\_verify} with boxed-answer fallback \\
    \bottomrule
  \end{tabular}
  \caption{Evaluation hyperparameters used in the experiments.}
  \label{tab:app_eval_hparams}
\end{table}

To test whether math-focused RLVR training induces out-of-domain (OOD) capability collapse, all Qwen3-4B checkpoints are additionally evaluated on MMLU-Pro~\cite{mmlupro}, whose task distribution differs from DAPO-Math-17K. Unlike the in-domain math benchmarks (32K-token cap in Table~\ref{tab:app_eval_hparams}), this OOD evaluation uses an 8K maximum generation token budget, with the same vLLM backend, non-thinking generation, temperature $0.7$, and 16 samples per problem. Table~\ref{tab:app_mmlu_pro_ood} reports Avg@16 and Pass@16 by subject in the same format as Table~\ref{tab:main_avg_pass16}; within each subject, the highest Avg and highest Pass across methods are in bold.

\begin{table}[h]
  \centering
  \setlength{\tabcolsep}{3.0pt}
  \renewcommand{\arraystretch}{1.06}
  \begin{adjustbox}{max width=0.96\textwidth}
  \begin{tabular}{@{}lcccccccccccc@{}}
    \toprule
    \textbf{Category}
    & \multicolumn{2}{c}{\textbf{Base}}
    & \multicolumn{2}{c}{\textbf{GRPO}}
    & \multicolumn{2}{c}{\textbf{OPSD}}
    & \multicolumn{2}{c}{\textbf{RLSD}}
    & \multicolumn{2}{c}{\textbf{RLRT}}
    & \multicolumn{2}{c}{\textbf{CAST}} \\
    \cmidrule(lr){2-3}
    \cmidrule(lr){4-5}
    \cmidrule(lr){6-7}
    \cmidrule(lr){8-9}
    \cmidrule(lr){10-11}
    \cmidrule(lr){12-13}
    & Avg & Pass & Avg & Pass & Avg & Pass & Avg & Pass & Avg & Pass & Avg & Pass \\
    \midrule
    Biology       & 79.02 & 93.03 & 79.31 & 93.03 & 78.71 & 92.61 & 79.15 & 93.03 & 79.38 & 93.03 & \textbf{80.92} & \textbf{94.28} \\
    Business      & 70.08 & 89.48 & 70.32 & 88.97 & 70.66 & 88.21 & 70.56 & 89.35 & 70.54 & \textbf{90.11} & \textbf{73.72} & 89.73 \\
    Chemistry     & 70.56 & 92.14 & 70.56 & 91.17 & 70.43 & 91.61 & 70.76 & 92.40 & 71.36 & 91.43 & \textbf{75.43} & \textbf{93.11} \\
    Comp.\ Sci.   & 67.79 & \textbf{88.78} & 68.63 & \textbf{88.78} & 68.05 & 87.56 & 68.57 & 87.56 & 68.99 & 87.56 & \textbf{70.90} & \textbf{88.78} \\
    Economics     & 72.07 & 88.39 & 72.03 & 88.98 & 71.91 & 88.27 & 72.04 & 88.51 & 71.90 & 87.56 & \textbf{73.88} & \textbf{89.45} \\
    Engineering   & 54.35 & 87.72 & 53.59 & 88.34 & 53.93 & 87.72 & 54.37 & 88.24 & 53.77 & 87.62 & \textbf{55.13} & \textbf{89.47} \\
    Health        & 58.21 & 78.00 & 57.93 & 79.58 & 57.84 & 78.61 & 57.64 & 78.85 & 57.94 & 78.85 & \textbf{59.59} & \textbf{79.34} \\
    History       & 47.85 & 68.77 & 47.28 & \textbf{70.87} & 47.90 & 69.55 & 47.47 & 70.08 & 47.77 & 70.34 & \textbf{48.16} & 70.08 \\
    Law           & 29.67 & 62.40 & 29.62 & 64.03 & 29.79 & 62.58 & 29.58 & 62.31 & 29.53 & 62.49 & \textbf{30.81} & \textbf{64.85} \\
    Math          & 80.46 & 94.08 & 80.83 & 93.93 & 80.67 & 94.30 & 80.75 & 94.45 & 81.30 & 94.30 & \textbf{85.21} & \textbf{95.19} \\
    Other         & 53.34 & 75.87 & 53.27 & 74.89 & 53.24 & 75.43 & 53.38 & 76.08 & 53.45 & 76.62 & \textbf{55.10} & \textbf{76.52} \\
    Philosophy    & 51.58 & 74.35 & 51.01 & 73.35 & 50.74 & 73.55 & 50.56 & 72.55 & 51.19 & 72.95 & \textbf{53.08} & \textbf{75.35} \\
    Physics       & 70.78 & 91.61 & 71.21 & 91.76 & 70.98 & 91.61 & 70.78 & 91.69 & 71.43 & 91.69 & \textbf{75.27} & \textbf{92.30} \\
    Psychology    & 65.97 & 82.08 & 66.03 & 81.70 & 66.15 & 81.33 & 65.88 & 81.08 & 66.17 & 81.08 & \textbf{66.72} & \textbf{82.33} \\
    \midrule
    \textbf{Average} & 63.16 & 84.37 & 63.18 & 84.52 & 63.13 & 84.15 & 63.17 & 84.41 & 63.41 & 84.32 & \textbf{65.74} & \textbf{85.43} \\
    \bottomrule
  \end{tabular}
  \end{adjustbox}
  \caption{MMLU-Pro OOD Avg@16 and Pass@16 (\%) on Qwen3-4B under an 8K generation token budget.}
  \label{tab:app_mmlu_pro_ood}
\end{table}

\noindent Under the 8K budget, baseline Avg@16 scores remain clustered near 63.1--63.4\%, indicating no severe OOD collapse relative to the base model; Pass@16 shows a similar pattern (84.1--84.5\%). CAST reaches 65.74\% Avg@16 and 85.43\% Pass@16 overall, improving over the strongest baselines while preserving the largest in-domain gains on mathematical reasoning benchmarks.

In addition to Avg@16 and Pass@16 used in the main result tables, Pass@1 is also reported from the first sampled generation for each problem, using the same evaluation outputs.

\begin{table}[h]
  \centering
  \begin{tabular}{@{}lccccc@{}}
    \toprule
    \textbf{Method} & \textbf{AIME24} & \textbf{AIME25} & \textbf{AIME26} & \textbf{MATH-500} & \textbf{HMMT25}\\
    \midrule
    \multicolumn{6}{@{}l}{\textbf{Qwen3-1.7B}} \\
    Base & 10.0 & 6.7 & 0.0 & 72.6 & 6.7 \\
    +GRPO & 10.0 & 6.7 & 6.7 & 75.6 & 3.3 \\
    +OPSD & 10.0 & 3.3 & 3.3 & 68.4 & 0.0 \\
    +GRPO+OPSD & 6.7 & 13.3 & \textbf{10.0} & 75.2 & 6.7 \\
    +RLSD & 3.3 & 10.0 & 6.7 & 75.2 & 10.0 \\
    +RLRT & 13.3 & 10.0 & 3.3 & 72.6 & 6.7 \\
    +CAST & \textbf{20.0} & \textbf{30.0} & \textbf{10.0} & \textbf{83.0} & \textbf{13.3} \\
    \midrule
    \multicolumn{6}{@{}l}{\textbf{Qwen3-4B}} \\
    Base & 20.0 & 20.0 & 23.3 & 82.8 & 10.0 \\
    +GRPO & 16.7 & 23.3 & 23.3 & 84.6 & 13.3 \\
    +OPSD & 6.7 & 10.0 & 10.0 & 72.4 & 0.0 \\
    +GRPO+OPSD & 23.3 & 20.0 & 16.7 & 83.8 & 13.3 \\
    +RLSD & 30.0 & 26.7 & 23.3 & 83.8 & 13.3 \\
    +RLRT & 20.0 & 23.3 & 20.0 & 82.8 & 13.3 \\
    +CAST & \textbf{46.7} & \textbf{30.0} & \textbf{40.0} & \textbf{89.2} & \textbf{20.0} \\
    \midrule
    \multicolumn{6}{@{}l}{\textbf{Qwen3-8B}} \\
    Base & 23.3 & 16.7 & 13.3 & 84.8 & 10.0 \\
    +GRPO & 23.3 & 23.3 & 16.7 & 83.2 & 6.7 \\
    +OPSD & 13.3 & 20.0 & 13.3 & 78.2 & 6.7 \\
    +GRPO+OPSD & 26.7 & 26.7 & 13.3 & 83.6 & 10.0 \\
    +RLSD & 23.3 & 23.3 & 20.0 & 84.0 & 6.7 \\
    +RLRT & 26.7 & 23.3 & 13.3 & 83.6 & 6.7 \\
    +CAST & \textbf{40.0} & \textbf{33.3} & \textbf{36.7} & \textbf{90.2} & \textbf{20.0} \\
    \bottomrule
  \end{tabular}
  \caption{Pass@1 performance (\%) across model scales.}
  \label{tab:4b_pass1}
\end{table}

Because CAST produces longer completions on average than the baselines, Table~\ref{tab:app_eval_output_length} reports evaluation mean output length. Values are dataset-averages of the per-benchmark mean number of generated tokens across all Pass@16 samples on AIME24, AIME25, AIME26, MATH-500, and HMMT25.

\begin{table}[h]
  \centering
  \setlength{\tabcolsep}{4.0pt}
  \renewcommand{\arraystretch}{1.04}
  \begin{adjustbox}{max width=0.96\textwidth}
  \begin{tabular}{@{}lccccccc@{}}
    \toprule
    \textbf{Model}
    & \textbf{Base}
    & \textbf{GRPO}
    & \textbf{OPSD}
    & \textbf{GRPO+OPSD}
    & \textbf{RLSD}
    & \textbf{RLRT}
    & \textbf{CAST} \\
    \midrule
    Qwen3-1.7B & 3131.9 & 3138.9 & 3408.1 & 3245.1 & 3439.6 & 3383.3 & 10085.0\\
    Qwen3-4B   & 5242.1 & 5259.2 & 4495.9 & 5173.2 & 5374.0 & 5921.5 & 9779.7\\
    Qwen3-8B   & 5064.4 & 5335.2 & 10137.2 & 5334.9 & 4965.3 & 5217.9 & 9505.9  \\
    \bottomrule
  \end{tabular}
  \end{adjustbox}
  \caption{Evaluation mean output length.}
  \label{tab:app_eval_output_length}
\end{table}

Using the same Qwen3-4B non-thinking evaluation outputs, this analysis additionally evaluates how performance changes under explicit generation-token budgets. Accuracy is recomputed at 1K, 2K, 4K, 8K, 16K, and 32K token budgets. The 32K column is the final evaluation setting used in the main Qwen3-4B comparison. This budget sweep covers AIME24, AIME25, AIME26, MATH-500, and HMMT25 for the methods with complete budget traces. Table~\ref{tab:app_token_budget_avg_pass16} reports both Avg@16 and Pass@16.

\begin{table}[h]
  \centering
  \setlength{\tabcolsep}{2.0pt}
  \renewcommand{\arraystretch}{1.03}
  \begin{adjustbox}{max width=\textwidth}
  \begin{tabular}{@{}lcccccccccccc@{}}
    \toprule
    \textbf{Token Length}
    & \multicolumn{2}{c}{\textbf{1K}}
    & \multicolumn{2}{c}{\textbf{2K}}
    & \multicolumn{2}{c}{\textbf{4K}}
    & \multicolumn{2}{c}{\textbf{8K}}
    & \multicolumn{2}{c}{\textbf{16K}}
    & \multicolumn{2}{c}{\textbf{32K}} \\
    \cmidrule(lr){2-3}
    \cmidrule(lr){4-5}
    \cmidrule(lr){6-7}
    \cmidrule(lr){8-9}
    \cmidrule(lr){10-11}
    \cmidrule(lr){12-13}
    & Avg & Pass & Avg & Pass & Avg & Pass & Avg & Pass & Avg & Pass & Avg & Pass \\
    \midrule
    \multicolumn{13}{@{}l}{\textbf{AIME24}} \\
    Base & 3.75 & 16.7 & 15.00 & \textbf{36.7} & 20.21 & 43.3 & 20.83 & 46.7 & 20.83 & 46.7 & 20.83 & 46.7 \\
    +GRPO & 4.17 & \textbf{20.0} & 15.63 & \textbf{36.7} & 20.63 & 43.3 & 21.04 & 43.3 & 21.04 & 43.3 & 21.04 & 43.3 \\
    +GRPO+OPSD & 4.17 & 16.7 & 16.67 & \textbf{36.7} & 22.08 & \textbf{53.3} & 22.50 & 53.3 & 22.50 & 53.3 & 22.50 & 53.3 \\
    +RLSD & \textbf{4.58} & \textbf{20.0} & \textbf{17.29} & \textbf{36.7} & 20.42 & 46.7 & 21.46 & 46.7 & 21.67 & 50.0 & 21.67 & 50.0 \\
    +RLRT & 3.33 & 13.3 & 15.63 & \textbf{36.7} & 20.42 & 43.3 & 21.25 & 50.0 & 21.67 & 53.3 & 21.67 & 53.3 \\
    +CAST & 2.92 & 6.7 & 15.42 & 30.0 & \textbf{26.04} & 50.0 & \textbf{36.67} & \textbf{70.0} & \textbf{40.83} & \textbf{76.7} & \textbf{41.25} & \textbf{76.7} \\
    \midrule
    \multicolumn{13}{@{}l}{\textbf{AIME25}} \\
    Base & \textbf{4.38} & 6.7 & 16.04 & 30.0 & 20.21 & \textbf{46.7} & 21.25 & 50.0 & 21.25 & 50.0 & 21.25 & 50.0 \\
    +GRPO & 4.17 & 6.7 & \textbf{18.33} & \textbf{33.3} & 21.67 & 43.3 & 22.50 & 50.0 & 22.50 & 50.0 & 22.50 & 50.0 \\
    +GRPO+OPSD & \textbf{4.38} & 6.7 & 17.71 & \textbf{33.3} & 20.83 & 40.0 & 21.67 & 46.7 & 21.67 & 46.7 & 21.67 & 46.7 \\
    +RLSD & 3.33 & 6.7 & 17.71 & \textbf{33.3} & 19.79 & 43.3 & 21.67 & 50.0 & 21.67 & 50.0 & 21.88 & 50.0 \\
    +RLRT & \textbf{4.38} & \textbf{10.0} & 16.04 & 30.0 & 20.83 & 43.3 & 22.08 & 46.7 & 22.08 & 46.7 & 22.08 & 46.7 \\
    +CAST & 1.25 & 6.7 & 17.29 & 30.0 & \textbf{22.92} & 40.0 & \textbf{29.58} & \textbf{53.3} & \textbf{35.00} & \textbf{70.0} & \textbf{35.83} & \textbf{70.0} \\
    \midrule
    \multicolumn{13}{@{}l}{\textbf{AIME26}} \\
    Base & 1.88 & 6.7 & 13.54 & 36.7 & 17.29 & \textbf{50.0} & 17.50 & 50.0 & 17.71 & 50.0 & 17.71 & 50.0 \\
    +GRPO & 1.88 & 6.7 & \textbf{15.42} & 33.3 & 19.38 & \textbf{50.0} & 20.00 & 53.3 & 20.00 & 53.3 & 20.00 & 53.3 \\
    +GRPO+OPSD & 1.46 & 6.7 & 13.54 & 33.3 & 16.67 & \textbf{50.0} & 17.08 & 50.0 & 17.29 & 50.0 & 17.29 & 50.0 \\
    +RLSD & 2.29 & 6.7 & 14.58 & 26.7 & 16.88 & 36.7 & 17.50 & 40.0 & 17.71 & 40.0 & 17.71 & 40.0 \\
    +RLRT & \textbf{2.50} & \textbf{10.0} & 14.58 & \textbf{40.0} & 17.92 & \textbf{50.0} & 19.58 & 60.0 & 20.00 & 63.3 & 20.00 & 63.3 \\
    +CAST & 0.21 & 3.3 & 11.67 & 26.7 & \textbf{24.79} & 43.3 & \textbf{32.50} & \textbf{63.3} & \textbf{35.83} & \textbf{70.0} & \textbf{36.25} & \textbf{70.0} \\
    \midrule
    \multicolumn{13}{@{}l}{\textbf{MATH-500}} \\
    Base & 70.24 & 83.8 & \textbf{81.11} & 93.0 & 83.66 & 95.6 & 83.89 & 96.0 & 83.91 & 96.0 & 83.91 & 96.0 \\
    +GRPO & 70.28 & \textbf{84.6} & 81.00 & \textbf{93.4} & 83.59 & 95.0 & 83.80 & 95.4 & 83.86 & 95.4 & 83.86 & 95.4 \\
    +GRPO+OPSD & \textbf{70.41} & 83.0 & 81.04 & \textbf{93.4} & 83.29 & 95.0 & 83.70 & 95.8 & 83.73 & 96.0 & 83.74 & 96.0 \\
    +RLSD & 70.28 & 83.8 & 81.09 & 93.2 & 83.69 & \textbf{96.0} & 84.06 & 96.2 & 84.08 & 96.2 & 84.08 & 96.2 \\
    +RLRT & 69.10 & 82.6 & 80.64 & 92.2 & 83.80 & 95.0 & 84.28 & 95.4 & 84.34 & 95.6 & 84.35 & 95.6 \\
    +CAST & 61.23 & 74.0 & 77.29 & 88.8 & \textbf{85.31} & 95.4 & \textbf{88.63} & \textbf{97.4} & \textbf{89.34} & \textbf{97.8} & \textbf{89.38} & \textbf{98.0} \\
    \midrule
    \multicolumn{13}{@{}l}{\textbf{HMMT25}} \\
    Base & 0.63 & \textbf{3.3} & 7.50 & 13.3 & 10.00 & 20.0 & 10.21 & 23.3 & 10.21 & 23.3 & 10.21 & 23.3 \\
    +GRPO & 0.42 & \textbf{3.3} & 8.75 & \textbf{16.7} & 11.67 & \textbf{30.0} & 12.08 & 30.0 & 12.08 & 30.0 & 12.08 & 30.0 \\
    +GRPO+OPSD & \textbf{1.04} & \textbf{3.3} & \textbf{9.58} & \textbf{16.7} & \textbf{12.50} & 20.0 & 12.92 & 23.3 & 12.92 & 23.3 & 12.92 & 23.3 \\
    +RLSD & 0.42 & \textbf{3.3} & \textbf{9.58} & \textbf{16.7} & 11.46 & 23.3 & 11.88 & 23.3 & 11.88 & 23.3 & 11.88 & 23.3 \\
    +RLRT & 0.21 & \textbf{3.3} & 8.54 & \textbf{16.7} & 12.29 & 20.0 & 12.92 & 26.7 & 12.92 & 26.7 & 12.92 & 26.7 \\
    +CAST & 0.00 & 0.0 & 6.88 & 10.0 & 12.29 & 20.0 & \textbf{17.50} & \textbf{36.7} & \textbf{21.46} & \textbf{50.0} & \textbf{21.67} & \textbf{50.0} \\
    \bottomrule
  \end{tabular}
  \end{adjustbox}
  \caption{Qwen3-4B performance (\%) under different generation-token budgets. Avg and Pass denote Avg@16 and Pass@16, respectively. The 32K columns are the final 32K evaluation results. Bold marks the best value within each dataset and token budget for each metric.}
  \label{tab:app_token_budget_avg_pass16}
\end{table}

\subsection{Seed Sensitivity for Qwen3-4B}

To check the sensitivity of the Qwen3-4B evaluation to sampling randomness, the main-comparison evaluations are repeated under two additional evaluation seeds, 0 and 1024, using the same evaluation protocol as in Table~\ref{tab:app_eval_hparams}. Table~\ref{tab:app_seed_4b} reports Avg@16 and Pass@16 on AIME24 and AIME26 while varying only the evaluation seed. CAST is also trained with training seeds 0 and 1024, and those checkpoints are evaluated with the default evaluation seed 42; these training-seed results are shown in Table~\ref{tab:app_train_seed_4b}.

\begin{table}[h]
  \centering
  \setlength{\tabcolsep}{3.2pt}
  \renewcommand{\arraystretch}{1.06}
  \begin{adjustbox}{max width=\textwidth}
  \begin{tabular}{@{}lcccccccc@{}}
    \toprule
    \textbf{Method}
    & \multicolumn{4}{c}{Eval seed 0}
    & \multicolumn{4}{c}{Eval seed 1024} \\
    \cmidrule(lr){2-5}
    \cmidrule(lr){6-9}
    & \multicolumn{2}{c}{AIME24}
    & \multicolumn{2}{c}{AIME26}
    & \multicolumn{2}{c}{AIME24}
    & \multicolumn{2}{c}{AIME26} \\
    \cmidrule(lr){2-3}
    \cmidrule(lr){4-5}
    \cmidrule(lr){6-7}
    \cmidrule(lr){8-9}
    & Avg & Pass & Avg & Pass & Avg & Pass & Avg & Pass \\
    \midrule
    Base         & 24.58 & 56.7 & 17.71 & 46.7 & 22.92 & 53.3 & 17.29 & 46.7 \\
    +GRPO        & 22.08 & 53.3 & 17.29 & 36.7 & 23.96 & 50.0 & 18.75 & 46.7 \\
    +OPSD        & 8.75 & 23.3 & 9.38 & 13.3 & 10.62 & 26.7 & 9.79 & 16.7 \\
    +GRPO+OPSD   & 23.96 & 53.3 & 18.75 & 50.0 & 22.71 & 50.0 & 17.50 & 36.7 \\
    +RLSD        & 23.12 & 56.7 & 16.46 & 36.7 & 22.92 & 46.7 & 17.29 & 46.7 \\
    +RLRT        & 22.50 & 50.0 & 19.38 & \textbf{63.3} & 23.75 & 53.3 & 17.71 & 53.3 \\
    +CAST & \textbf{37.71} & \textbf{80.0} & \textbf{34.38} & \textbf{63.3} & \textbf{36.25} & \textbf{76.7} & \textbf{32.71} & \textbf{70.0} \\
    \bottomrule
  \end{tabular}
  \end{adjustbox}
  \caption{Qwen3-4B evaluation-seed sweep on AIME24 and AIME26. Avg and Pass denote Avg@16 and Pass@16, respectively.}
  \label{tab:app_seed_4b}
\end{table}

\begin{table}[h]
  \centering
  \setlength{\tabcolsep}{4.0pt}
  \renewcommand{\arraystretch}{1.06}
  \begin{adjustbox}{max width=0.82\textwidth}
  \begin{tabular}{@{}lcccc@{}}
    \toprule
    \textbf{CAST training seed}
    & \multicolumn{2}{c}{AIME24}
    & \multicolumn{2}{c}{AIME26} \\
    \cmidrule(lr){2-3}
    \cmidrule(lr){4-5}
    & Avg & Pass & Avg & Pass \\
    \midrule
    Default & \textbf{41.25} & 76.7 & 36.25 & \textbf{70.0} \\
    0 & 38.54 & \textbf{80.0} & \textbf{37.71} & \textbf{70.0} \\
    1024 & 40.00 & \textbf{80.0} & 37.08 & \textbf{70.0} \\
    \bottomrule
  \end{tabular}
  \end{adjustbox}
  \caption{Qwen3-4B CAST training-seed sweep evaluated with the default evaluation seed 42. Avg and Pass denote Avg@16 and Pass@16, respectively.}
  \label{tab:app_train_seed_4b}
\end{table}

\subsection{Additional Explorations}
\label{app:additional_explorations}

Table~\ref{tab:exploration-avg-pass16} summarizes the exploratory path from RLSD to CAST. All results are evaluated with 16 samples under the non-thinking 32K setting. The row labeled RLSD (zero-variance updates) denotes an exploratory variant that also updates zero-variance all-correct and all-wrong groups. Starting from RLSD, the variants progressively add zero-variance updates, strict group handling, sign flipping, removal of the ground-truth teacher reference, non-decayed token-gap shaping, and finally the incorrect-path positive flip used by CAST.

\begin{table}[h]
\centering
\setlength{\tabcolsep}{3pt}
\renewcommand{\arraystretch}{1.12}
\begin{adjustbox}{max width=\textwidth}
\begin{tabular}{lcccccccccc}
\toprule
\textbf{Method}
& \multicolumn{2}{c}{AIME24}
& \multicolumn{2}{c}{AIME25}
& \multicolumn{2}{c}{AIME26}
& \multicolumn{2}{c}{MATH-500}
& \multicolumn{2}{c}{HMMT25} \\
\cmidrule(lr){2-3}
\cmidrule(lr){4-5}
\cmidrule(lr){6-7}
\cmidrule(lr){8-9}
\cmidrule(lr){10-11}
& Avg & Pass & Avg & Pass & Avg & Pass & Avg & Pass & Avg & Pass \\
\midrule
RLSD & 21.67 & 50.00 & 21.88 & 50.00 & 17.71 & 40.00 & 84.08 & 96.20 & 11.88 & 23.33 \\
RLSD (zero-variance updates) & 20.21 & 43.33 & 20.21 & 40.00 & 17.92 & 53.33 & 83.71 & 95.80 & 11.25 & 26.67 \\
Strict RLSD & 23.12 & 46.67 & 22.71 & 43.33 & 18.75 & 46.67 & 83.89 & 95.60 & 13.54 & 30.00 \\
Flip RLSD & 29.58 & 73.33 & 23.54 & 56.67 & 22.50 & 50.00 & 86.31 & 96.60 & 15.21 & 33.33 \\
Flip RLSD (answer-free) & 33.33 & 63.33 & 25.83 & 53.33 & 23.96 & \textbf{70.00} & 86.85 & 96.00 & 16.25 & 33.33 \\
Flip RLSD (answer-free, no decay) & 28.75 & 56.67 & 25.00 & 53.33 & 24.58 & 60.00 & 86.78 & 96.20 & 14.38 & 36.67 \\
CAST & \textbf{41.25} & \textbf{76.67} & \textbf{35.83} & \textbf{70.00} & \textbf{36.25} & \textbf{70.00} & \textbf{89.38} & \textbf{98.00} & \textbf{21.67} & \textbf{50.00} \\
\bottomrule
\end{tabular}
\end{adjustbox}
\caption{Avg@16 and Pass@16 across RLSD exploration variants.}
\label{tab:exploration-avg-pass16}
\end{table}

\subsubsection{RLSD}

The RLSD baseline applies token-level shaping on top of the sequence-level GRPO advantage. It improves over sparse sequence-level feedback by using the teacher--student token log-probability gap to modulate each token's advantage. However, this formulation still depends on the original group-relative advantage, which becomes weak when all sampled responses for a prompt are correct or all are incorrect.

\subsubsection{RLSD with Zero-Variance Updates}

The RLSD (zero-variance updates) variant additionally applies RLSD-style updates to all-correct and all-wrong groups. Because this changes the handling of zero-variance groups while retaining the RLSD shaping rule, it is treated as an exploratory variant rather than as the main RLSD baseline.

\subsubsection{Strict RLSD}

Strict RLSD addresses this issue by splitting groups into mixed, all-correct, and all-wrong cases. Mixed groups keep the original GRPO advantage, while all-correct and all-wrong groups receive explicit fixed base advantages. This gives nonzero training signal even when group-relative normalization would otherwise collapse.

\subsubsection{Flip RLSD}

This variant introduces local advantage sign reversal. A token from a positive-advantage rollout can still receive negative credit if the teacher--student gap is negative. This avoids uniformly reinforcing every token in a globally successful response and makes the token-level feedback more selective.

\subsubsection{Flip RLSD without Ground-Truth Teacher Reference}

This variant switches to answer-free self-teacher scoring by removing the reference solution from the teacher prompt, so the token-level signal depends only on the problem and the student response. This variant tests whether RLSD can provide useful token-level supervision without access to the ground-truth solution or reference solution.

\subsubsection{Flip RLSD without Ground-Truth Reference and without Decay}

This variant keeps the token-gap shaping active throughout training instead of annealing it away. It treats the teacher--student token gap as a persistent training signal rather than only as an early-stage regularizer.

\subsubsection{CAST with Incorrect-Path Positive Flipping}

CAST adds the final bidirectional sign-reversal rule: teacher-negative positive-base tokens receive negative reversal, and teacher-positive negative-base tokens receive bounded positive reversal under the positive-base clipping cap. This is the configuration used in the main results.

\subsection{Full-Trajectory Token Map of Teacher-Positive and Teacher-Negative Signals}
\label{app:token_shift_visualization}

\providecommand{\tokPW}[1]{\begingroup\setlength{\fboxsep}{0.18pt}\colorbox{green!10}{\strut\ttfamily #1}\endgroup}
\providecommand{\tokPM}[1]{\begingroup\setlength{\fboxsep}{0.18pt}\colorbox{green!22}{\strut\ttfamily #1}\endgroup}
\providecommand{\tokPS}[1]{\begingroup\setlength{\fboxsep}{0.18pt}\colorbox{green!40}{\strut\ttfamily #1}\endgroup}
\providecommand{\tokLW}[1]{\begingroup\setlength{\fboxsep}{0.18pt}\colorbox{red!10}{\strut\ttfamily #1}\endgroup}
\providecommand{\tokLM}[1]{\begingroup\setlength{\fboxsep}{0.18pt}\colorbox{red!22}{\strut\ttfamily #1}\endgroup}
\providecommand{\tokLS}[1]{\begingroup\setlength{\fboxsep}{0.18pt}\colorbox{red!40}{\strut\ttfamily #1}\endgroup}
\providecommand{\tokSame}[1]{\begingroup\setlength{\fboxsep}{0.18pt}\colorbox{gray!12}{\strut\ttfamily #1}\endgroup}

To make the token-level motivation concrete, Figures~\ref{fig:full_token_shift_map} and~\ref{fig:full_token_shift_map_wrong} visualize every token in two representative OPSD diagnostic trajectories. The diagnostic setup uses an OPSD-style teacher context with a reference answer; these figures illustrate teacher--student gap behavior only and do not depict answer-free self-teacher scoring as used by CAST at training time. For each generated token $y_t$, define
\[
\Delta_t =
\log \pi_{\mathrm{teacher}}(y_t \mid \mathrm{context}, y_{<t})
-
\log \pi_{\mathrm{student}}(y_t \mid x,y_{<t}).
\]
Green tokens have $\Delta_t>0$ (teacher-positive); red tokens have $\Delta_t<0$ (teacher-negative). Darker colors indicate larger $|\Delta_t|$. Blank colored cells correspond to whitespace tokens, and \texttt{\textbackslash{}n} marks newline tokens.

\paragraph{Correct rollout.}
Figure~\ref{fig:full_token_shift_map} shows a compact but complete correct rollout with 537 generated tokens. The reference answer is 13, and the generated answer is ``Answer: 13<|im\_end|>'', so the trajectory receives verifier reward 1.0. In this trajectory, 306 tokens are teacher-positive, 215 are teacher-negative, and 16 are nearly unchanged; the total teacher-positive weight is 53.89, while the total teacher-negative weight is 104.96. Thus, even a correct rollout contains many local tokens that the teacher would down-weight rather than reinforce.

\begin{figure}[h]
\centering
\setlength{\fboxrule}{0pt}
\begin{minipage}{0.99\linewidth}
\raggedright
{\scriptsize\textbf{Problem.} An ordinary $6$-sided die has a number on each face from $1$ to $6$; each number appears on one face. How many ways can two faces be painted blue so that the product of the numbers on the painted faces is not equal to $6$?\par}
\vspace{0.35em}
\fontsize{5.8pt}{6.2pt}\selectfont
\tokLS{We} \tokLS{\hspace{0.28em}are} \tokLM{\hspace{0.28em}given} \tokPM{\hspace{0.28em}an} \tokLW{\hspace{0.28em}ordinary} \tokLW{\hspace{0.28em}} \tokLW{6} \tokPW{-sided} \tokPW{\hspace{0.28em}die}
\tokLM{,} \tokLM{\hspace{0.28em}with} \tokSame{\hspace{0.28em}faces} \tokPM{\hspace{0.28em}numbered} \tokPS{\hspace{0.28em}from} \tokLW{\hspace{0.28em}} \tokLW{1} \tokPW{\hspace{0.28em}to} \tokLW{\hspace{0.28em}}
\tokLW{6} \tokPM{.} \tokLM{\hspace{0.28em}We} \tokLM{\hspace{0.28em}want} \tokLW{\hspace{0.28em}to} \tokLM{\hspace{0.28em}count} \tokPM{\hspace{0.28em}the} \tokLW{\hspace{0.28em}number} \tokLW{\hspace{0.28em}of}
\tokPW{\hspace{0.28em}ways} \tokPW{\hspace{0.28em}to} \tokPW{\hspace{0.28em}paint} \tokLS{\hspace{0.28em}**} \tokPW{two} \tokLM{\hspace{0.28em}faces} \tokPS{\hspace{0.28em}blue} \tokPW{**} \tokPW{\hspace{0.28em}such}
\tokPW{\hspace{0.28em}that} \tokLW{\hspace{0.28em}the} \tokPW{\hspace{0.28em}**} \tokPW{product} \tokLW{\hspace{0.28em}of} \tokLW{\hspace{0.28em}the} \tokLW{\hspace{0.28em}numbers} \tokLW{\hspace{0.28em}on} \tokPW{\hspace{0.28em}the}
\tokPM{\hspace{0.28em}painted} \tokLW{\hspace{0.28em}faces} \tokLM{\hspace{0.28em}is} \tokPM{\hspace{0.28em}not} \tokLW{\hspace{0.28em}equal} \tokPW{\hspace{0.28em}to} \tokPW{\hspace{0.28em}} \tokPW{6} \tokPW{**}
\tokLM{.{\color{gray!70}\textbackslash{}n}{\color{gray!70}\textbackslash{}n}} \tokLW{---{\color{gray!70}\textbackslash{}n}{\color{gray!70}\textbackslash{}n}} \tokLW{\#\#\#} \tokLM{\hspace{0.28em}Step} \tokPW{\hspace{0.28em}} \tokLW{1} \tokLW{:} \tokPW{\hspace{0.28em}Total} \tokLW{\hspace{0.28em}number}
\tokPW{\hspace{0.28em}of} \tokLW{\hspace{0.28em}ways} \tokPW{\hspace{0.28em}to} \tokLM{\hspace{0.28em}choose} \tokLM{\hspace{0.28em}} \tokSame{2} \tokLW{\hspace{0.28em}faces} \tokPW{\hspace{0.28em}to} \tokPW{\hspace{0.28em}paint}
\tokPW{\hspace{0.28em}blue} \tokPW{{\color{gray!70}\textbackslash{}n}{\color{gray!70}\textbackslash{}n}} \tokLM{There} \tokLW{\hspace{0.28em}are} \tokLW{\hspace{0.28em}} \tokLW{6} \tokLW{\hspace{0.28em}faces} \tokLW{\hspace{0.28em}on} \tokPW{\hspace{0.28em}the}
\tokPW{\hspace{0.28em}die} \tokPM{,} \tokLW{\hspace{0.28em}and} \tokLW{\hspace{0.28em}we} \tokLW{\hspace{0.28em}are} \tokLM{\hspace{0.28em}choosing} \tokLW{\hspace{0.28em}} \tokLW{2} \tokLM{\hspace{0.28em}of}
\tokPW{\hspace{0.28em}them} \tokLM{\hspace{0.28em}to} \tokPW{\hspace{0.28em}paint} \tokLW{\hspace{0.28em}blue} \tokPW{.} \tokLM{\hspace{0.28em}The} \tokLW{\hspace{0.28em}number} \tokPW{\hspace{0.28em}of} \tokLM{\hspace{0.28em}ways}
\tokLW{\hspace{0.28em}to} \tokLM{\hspace{0.28em}choose} \tokPW{\hspace{0.28em}} \tokPW{2} \tokLW{\hspace{0.28em}faces} \tokPM{\hspace{0.28em}out} \tokPW{\hspace{0.28em}of} \tokPW{\hspace{0.28em}} \tokPW{6}
\tokLW{\hspace{0.28em}is} \tokLM{:{\color{gray!70}\textbackslash{}n}{\color{gray!70}\textbackslash{}n}} \tokPW{\$\$} \tokPW{{\color{gray!70}\textbackslash{}n}} \tokPW{\textbackslash{}} \tokLW{bin} \tokPW{om} \tokPW{\{} \tokPW{6}
\tokPW{\}\{} \tokPW{2} \tokLW{\}} \tokPW{\hspace{0.28em}=} \tokLM{\hspace{0.28em}} \tokLW{1} \tokPW{5} \tokPW{{\color{gray!70}\textbackslash{}n}} \tokPW{\$\$}
\tokPW{{\color{gray!70}\textbackslash{}n}{\color{gray!70}\textbackslash{}n}} \tokLW{So} \tokLM{\hspace{0.28em}there} \tokPW{\hspace{0.28em}are} \tokLM{\hspace{0.28em}} \tokLW{1} \tokPW{5} \tokPM{\hspace{0.28em}total} \tokLM{\hspace{0.28em}ways}
\tokLW{\hspace{0.28em}to} \tokLM{\hspace{0.28em}choose} \tokLM{\hspace{0.28em}} \tokLW{2} \tokLW{\hspace{0.28em}faces} \tokLM{\hspace{0.28em}to} \tokPW{\hspace{0.28em}paint} \tokLW{\hspace{0.28em}blue} \tokLM{.{\color{gray!70}\textbackslash{}n}{\color{gray!70}\textbackslash{}n}}
\tokPW{---{\color{gray!70}\textbackslash{}n}{\color{gray!70}\textbackslash{}n}} \tokLW{\#\#\#} \tokSame{\hspace{0.28em}Step} \tokSame{\hspace{0.28em}} \tokPW{2} \tokSame{:} \tokLW{\hspace{0.28em}Count} \tokPW{\hspace{0.28em}the} \tokPW{\hspace{0.28em}number}
\tokSame{\hspace{0.28em}of} \tokLM{\hspace{0.28em}pairs} \tokPM{\hspace{0.28em}whose} \tokLW{\hspace{0.28em}product} \tokLW{\hspace{0.28em}is} \tokLM{\hspace{0.28em}} \tokPW{6} \tokLW{{\color{gray!70}\textbackslash{}n}{\color{gray!70}\textbackslash{}n}} \tokLW{We}
\tokLM{\hspace{0.28em}now} \tokPM{\hspace{0.28em}need} \tokPW{\hspace{0.28em}to} \tokPW{\hspace{0.28em}find} \tokPM{\hspace{0.28em}how} \tokPW{\hspace{0.28em}many} \tokPW{\hspace{0.28em}of} \tokPW{\hspace{0.28em}these} \tokLW{\hspace{0.28em}}
\tokLW{1} \tokPW{5} \tokPW{\hspace{0.28em}pairs} \tokPM{\hspace{0.28em}have} \tokLW{\hspace{0.28em}a} \tokPS{\hspace{0.28em}**} \tokPW{product} \tokLM{\hspace{0.28em}of} \tokLW{\hspace{0.28em}}
\tokLW{6} \tokLM{**} \tokPW{.} \tokPS{\hspace{0.28em}These} \tokLM{\hspace{0.28em}are} \tokLW{\hspace{0.28em}the} \tokLW{\hspace{0.28em}pairs} \tokPM{\hspace{0.28em}we} \tokLM{\hspace{0.28em}**}
\tokPW{do} \tokPM{\hspace{0.28em}not} \tokPS{**} \tokPW{\hspace{0.28em}want} \tokLM{\hspace{0.28em}to} \tokLW{\hspace{0.28em}count} \tokPW{.{\color{gray!70}\textbackslash{}n}{\color{gray!70}\textbackslash{}n}} \tokPM{We} \tokLM{\hspace{0.28em}list}
\tokLW{\hspace{0.28em}all} \tokLW{\hspace{0.28em}pairs} \tokLW{\hspace{0.28em}of} \tokPM{\hspace{0.28em}numbers} \tokLW{\hspace{0.28em}from} \tokLW{\hspace{0.28em}} \tokLW{1} \tokPW{\hspace{0.28em}to} \tokPW{\hspace{0.28em}}
\tokLW{6} \tokLW{\hspace{0.28em}and} \tokPM{\hspace{0.28em}find} \tokLW{\hspace{0.28em}which} \tokPW{\hspace{0.28em}pairs} \tokPW{\hspace{0.28em}have} \tokPW{\hspace{0.28em}a} \tokPW{\hspace{0.28em}product} \tokLW{\hspace{0.28em}of}
\tokPW{\hspace{0.28em}} \tokSame{6} \tokLW{:{\color{gray!70}\textbackslash{}n}{\color{gray!70}\textbackslash{}n}} \tokPW{-} \tokPM{\hspace{0.28em}\$} \tokPM{1} \tokPW{\hspace{0.28em}\textbackslash{}} \tokPW{times} \tokPW{\hspace{0.28em}}
\tokLW{6} \tokPW{\hspace{0.28em}=} \tokLW{\hspace{0.28em}} \tokPW{6} \tokPM{\${\color{gray!70}\textbackslash{}n}} \tokPW{-} \tokPW{\hspace{0.28em}\$} \tokPW{2} \tokPW{\hspace{0.28em}\textbackslash{}}
\tokSame{times} \tokSame{\hspace{0.28em}} \tokPW{3} \tokPW{\hspace{0.28em}=} \tokSame{\hspace{0.28em}} \tokPW{6} \tokPM{\${\color{gray!70}\textbackslash{}n}{\color{gray!70}\textbackslash{}n}} \tokLM{So} \tokPW{\hspace{0.28em}the}
\tokLM{\hspace{0.28em}pairs} \tokPS{\hspace{0.28em}of} \tokPM{\hspace{0.28em}numbers} \tokLW{\hspace{0.28em}that} \tokLW{\hspace{0.28em}multiply} \tokLW{\hspace{0.28em}to} \tokPW{\hspace{0.28em}} \tokPW{6} \tokPW{\hspace{0.28em}are}
\tokLM{:{\color{gray!70}\textbackslash{}n}} \tokLW{-} \tokLW{\hspace{0.28em}(} \tokPW{1} \tokPW{,} \tokPW{\hspace{0.28em}} \tokLW{6} \tokPW{){\color{gray!70}\textbackslash{}n}} \tokSame{-}
\tokPW{\hspace{0.28em}(} \tokPW{2} \tokPW{,} \tokPW{\hspace{0.28em}} \tokPW{3} \tokPW{){\color{gray!70}\textbackslash{}n}{\color{gray!70}\textbackslash{}n}} \tokLM{These} \tokPW{\hspace{0.28em}are} \tokPW{\hspace{0.28em}the}
\tokLW{\hspace{0.28em}only} \tokLM{\hspace{0.28em}two} \tokPM{\hspace{0.28em}such} \tokLW{\hspace{0.28em}pairs} \tokLW{.{\color{gray!70}\textbackslash{}n}{\color{gray!70}\textbackslash{}n}} \tokPM{Now} \tokPM{,} \tokPM{\hspace{0.28em}we} \tokPM{\hspace{0.28em}need}
\tokPW{\hspace{0.28em}to} \tokLM{\hspace{0.28em}count} \tokLW{\hspace{0.28em}how} \tokPW{\hspace{0.28em}many} \tokLS{\hspace{0.28em}of} \tokLW{\hspace{0.28em}the} \tokLM{\hspace{0.28em}} \tokLW{1} \tokPW{5}
\tokPM{\hspace{0.28em}total} \tokPM{\hspace{0.28em}face} \tokLM{\hspace{0.28em}pairs} \tokPM{\hspace{0.28em}correspond} \tokLW{\hspace{0.28em}to} \tokLW{\hspace{0.28em}these} \tokPM{\hspace{0.28em}number} \tokLW{\hspace{0.28em}pairs} \tokPW{.{\color{gray!70}\textbackslash{}n}{\color{gray!70}\textbackslash{}n}}
\tokPS{On} \tokPM{\hspace{0.28em}a} \tokLM{\hspace{0.28em}die} \tokLW{,} \tokPM{\hspace{0.28em}the} \tokPM{\hspace{0.28em}numbers} \tokLM{\hspace{0.28em}on} \tokLM{\hspace{0.28em}the} \tokLW{\hspace{0.28em}faces}
\tokLW{\hspace{0.28em}are} \tokPM{\hspace{0.28em}**} \tokPS{1} \tokLM{\hspace{0.28em}through} \tokPW{\hspace{0.28em}} \tokPW{6} \tokPM{**,} \tokPS{\hspace{0.28em}but} \tokPM{\hspace{0.28em}the}
\tokPM{\hspace{0.28em}**} \tokPW{actual} \tokLS{\hspace{0.28em}faces} \tokLW{**} \tokLW{\hspace{0.28em}are} \tokPS{\hspace{0.28em}fixed} \tokPM{\hspace{0.28em}in} \tokPW{\hspace{0.28em}a} \tokPS{\hspace{0.28em}}
\tokPW{3} \tokPW{D} \tokLM{\hspace{0.28em}object} \tokLW{.} \tokPM{\hspace{0.28em}However} \tokPW{,} \tokLW{\hspace{0.28em}the} \tokLM{\hspace{0.28em}**} \tokPS{problem}
\tokPW{\hspace{0.28em}doesn} \tokLW{'t} \tokLM{\hspace{0.28em}specify} \tokPM{\hspace{0.28em}any} \tokLS{\hspace{0.28em}restrictions} \tokLW{\hspace{0.28em}on} \tokPM{\hspace{0.28em}the} \tokLM{\hspace{0.28em}arrangement} \tokLW{\hspace{0.28em}of}
\tokPM{\hspace{0.28em}the} \tokLM{\hspace{0.28em}numbers} \tokPM{**} \tokPM{\hspace{0.28em}on} \tokLW{\hspace{0.28em}the} \tokPW{\hspace{0.28em}die} \tokLM{,} \tokLM{\hspace{0.28em}so} \tokPW{\hspace{0.28em}we}
\tokLM{\hspace{0.28em}assume} \tokPM{\hspace{0.28em}that} \tokLM{\hspace{0.28em}each} \tokPM{\hspace{0.28em}number} \tokLM{\hspace{0.28em}from} \tokPW{\hspace{0.28em}} \tokLW{1} \tokPW{\hspace{0.28em}to} \tokPW{\hspace{0.28em}}
\tokSame{6} \tokPW{\hspace{0.28em}is} \tokPW{\hspace{0.28em}on} \tokPW{\hspace{0.28em}a} \tokLM{\hspace{0.28em}different} \tokPW{\hspace{0.28em}face} \tokLW{.{\color{gray!70}\textbackslash{}n}{\color{gray!70}\textbackslash{}n}} \tokPM{Thus} \tokPW{,}
\tokLW{\hspace{0.28em}the} \tokPM{\hspace{0.28em}number} \tokPW{\hspace{0.28em}of} \tokPW{\hspace{0.28em}ways} \tokPW{\hspace{0.28em}to} \tokPM{\hspace{0.28em}choose} \tokLM{\hspace{0.28em}a} \tokLW{\hspace{0.28em}pair} \tokLW{\hspace{0.28em}of}
\tokPM{\hspace{0.28em}faces} \tokLM{\hspace{0.28em}that} \tokPS{\hspace{0.28em}correspond} \tokPW{\hspace{0.28em}to} \tokPM{\hspace{0.28em}the} \tokLM{\hspace{0.28em}number} \tokLM{\hspace{0.28em}pairs} \tokPW{\hspace{0.28em}(} \tokPW{1}
\tokLW{,} \tokLM{\hspace{0.28em}} \tokPW{6} \tokPW{)} \tokPM{\hspace{0.28em}and} \tokPW{\hspace{0.28em}(} \tokPW{2} \tokPW{,} \tokPW{\hspace{0.28em}}
\tokPW{3} \tokPW{)} \tokPW{\hspace{0.28em}is} \tokLW{:{\color{gray!70}\textbackslash{}n}{\color{gray!70}\textbackslash{}n}} \tokLW{-} \tokLM{\hspace{0.28em}(} \tokLW{1} \tokLW{,} \tokPW{\hspace{0.28em}}
\tokPW{6} \tokPM{):} \tokPM{\hspace{0.28em}} \tokPM{1} \tokPM{\hspace{0.28em}way} \tokLM{\hspace{0.28em}(} \tokPS{face} \tokLM{\hspace{0.28em}} \tokPW{1}
\tokPW{\hspace{0.28em}and} \tokPW{\hspace{0.28em}face} \tokPW{\hspace{0.28em}} \tokPW{6} \tokPW{){\color{gray!70}\textbackslash{}n}} \tokLW{-} \tokPW{\hspace{0.28em}(} \tokPW{2} \tokPW{,}
\tokPW{\hspace{0.28em}} \tokPW{3} \tokPW{):} \tokPW{\hspace{0.28em}} \tokPW{1} \tokPW{\hspace{0.28em}way} \tokPW{\hspace{0.28em}(} \tokPW{face} \tokPW{\hspace{0.28em}}
\tokPW{2} \tokPW{\hspace{0.28em}and} \tokPW{\hspace{0.28em}face} \tokSame{\hspace{0.28em}} \tokLW{3} \tokPW{){\color{gray!70}\textbackslash{}n}{\color{gray!70}\textbackslash{}n}} \tokLM{So} \tokPM{\hspace{0.28em}there} \tokPW{\hspace{0.28em}are}
\tokPM{\hspace{0.28em}**} \tokPW{2} \tokPM{**} \tokLM{\hspace{0.28em}such} \tokPW{\hspace{0.28em}pairs} \tokLW{\hspace{0.28em}that} \tokPM{\hspace{0.28em}result} \tokPW{\hspace{0.28em}in} \tokPW{\hspace{0.28em}a}
\tokPW{\hspace{0.28em}product} \tokPW{\hspace{0.28em}of} \tokPW{\hspace{0.28em}} \tokPW{6} \tokPW{.{\color{gray!70}\textbackslash{}n}{\color{gray!70}\textbackslash{}n}} \tokPW{---{\color{gray!70}\textbackslash{}n}{\color{gray!70}\textbackslash{}n}} \tokLW{\#\#\#} \tokPW{\hspace{0.28em}Step} \tokSame{\hspace{0.28em}}
\tokPW{3} \tokSame{:} \tokPW{\hspace{0.28em}Subtract} \tokPM{\hspace{0.28em}the} \tokPM{\hspace{0.28em}invalid} \tokPW{\hspace{0.28em}pairs} \tokLM{{\color{gray!70}\textbackslash{}n}{\color{gray!70}\textbackslash{}n}} \tokPM{We} \tokPM{\hspace{0.28em}have}
\tokLM{\hspace{0.28em}} \tokLW{1} \tokPW{5} \tokPW{\hspace{0.28em}total} \tokPM{\hspace{0.28em}ways} \tokLW{\hspace{0.28em}to} \tokPW{\hspace{0.28em}choose} \tokLW{\hspace{0.28em}} \tokPW{2}
\tokLW{\hspace{0.28em}faces} \tokPM{,} \tokPW{\hspace{0.28em}and} \tokPM{\hspace{0.28em}} \tokPW{2} \tokPW{\hspace{0.28em}of} \tokPM{\hspace{0.28em}them} \tokPM{\hspace{0.28em}result} \tokPW{\hspace{0.28em}in}
\tokPW{\hspace{0.28em}a} \tokPW{\hspace{0.28em}product} \tokLW{\hspace{0.28em}of} \tokPW{\hspace{0.28em}} \tokPW{6} \tokLW{.} \tokPS{\hspace{0.28em}Therefore} \tokPW{,} \tokPW{\hspace{0.28em}the}
\tokPW{\hspace{0.28em}number} \tokPW{\hspace{0.28em}of} \tokPW{\hspace{0.28em}valid} \tokPM{\hspace{0.28em}ways} \tokLM{\hspace{0.28em}(} \tokLM{where} \tokPW{\hspace{0.28em}the} \tokPW{\hspace{0.28em}product} \tokLW{\hspace{0.28em}is}
\tokLM{\hspace{0.28em}**} \tokPW{not} \tokPW{**} \tokLW{\hspace{0.28em}} \tokPW{6} \tokPW{)} \tokPW{\hspace{0.28em}is} \tokPW{:{\color{gray!70}\textbackslash{}n}{\color{gray!70}\textbackslash{}n}} \tokPW{\$\$}
\tokPW{{\color{gray!70}\textbackslash{}n}} \tokPW{1} \tokPW{5} \tokPW{\hspace{0.28em}-} \tokSame{\hspace{0.28em}} \tokPW{2} \tokPW{\hspace{0.28em}=} \tokLM{\hspace{0.28em}} \tokPW{1}
\tokPW{3} \tokPW{{\color{gray!70}\textbackslash{}n}} \tokLW{\$\$} \tokPW{{\color{gray!70}\textbackslash{}n}{\color{gray!70}\textbackslash{}n}} \tokLW{---{\color{gray!70}\textbackslash{}n}{\color{gray!70}\textbackslash{}n}} \tokPW{\#\#\#} \tokLM{\hspace{0.28em}Final} \tokPW{\hspace{0.28em}Answer} \tokLW{:{\color{gray!70}\textbackslash{}n}{\color{gray!70}\textbackslash{}n}}
\tokLS{Answer} \tokLW{:} \tokLS{\hspace{0.28em}} \tokLS{1} \tokLW{3} \tokLS{<|im\_end|>}
\end{minipage}
\caption{Full-token map of teacher-positive and teacher-negative signals for one correct trajectory. The rollout reaches the correct answer, but teacher-negative tokens appear throughout the reasoning trace and are especially strong around some answer-template tokens. This supports the CAST design choice of allowing token-level credit correction inside globally correct rollouts rather than assigning uniformly positive advantages to every token.}
\label{fig:full_token_shift_map}
\end{figure}

\paragraph{Incorrect rollout.}
Figure~\ref{fig:full_token_shift_map_wrong} shows a complementary incorrect trajectory: the reference answer is 2040, while the generated answer is ``120''. It contains 561 generated tokens, with 291 teacher-positive tokens, 256 teacher-negative tokens, and 14 nearly unchanged tokens. The teacher-positive weight is 45.67, while the teacher-negative weight is 85.75. Even an incorrect rollout can contain many locally teacher-positive tokens, so teacher-positive signals alone should not be treated as positive supervision without verifier-grounded trajectory direction.

\begin{figure}[H]
\centering
\setlength{\fboxrule}{0pt}
\begin{minipage}{0.99\linewidth}
\raggedright
{\scriptsize\textbf{Problem.} There are $5$ students on a team for a math competition. The competition has $5$ subject tests. Each student on the team must choose $2$ distinct tests, and each test must be taken by exactly two people. In how many ways can this be done?\par}
\vspace{0.35em}
\fontsize{5.8pt}{6.2pt}\selectfont
\tokLS{We} \tokLS{\hspace{0.28em}are} \tokLM{\hspace{0.28em}given} \tokLS{\hspace{0.28em}the} \tokLS{\hspace{0.28em}following} \tokLS{\hspace{0.28em}scenario} \tokLS{:{\color{gray!70}\textbackslash{}n}{\color{gray!70}\textbackslash{}n}} \tokLS{-} \tokLM{\hspace{0.28em}There}
\tokLW{\hspace{0.28em}are} \tokLS{\hspace{0.28em}**} \tokLW{5} \tokLW{\hspace{0.28em}students} \tokPW{**} \tokPM{\hspace{0.28em}on} \tokLW{\hspace{0.28em}a} \tokPW{\hspace{0.28em}team} \tokLM{.{\color{gray!70}\textbackslash{}n}}
\tokLW{-} \tokLM{\hspace{0.28em}There} \tokLW{\hspace{0.28em}are} \tokLW{\hspace{0.28em}**} \tokLW{5} \tokLM{\hspace{0.28em}subject} \tokLW{\hspace{0.28em}tests} \tokPW{**} \tokPW{.{\color{gray!70}\textbackslash{}n}}
\tokLW{-} \tokPW{\hspace{0.28em}Each} \tokPW{\hspace{0.28em}student} \tokPW{\hspace{0.28em}must} \tokPW{\hspace{0.28em}choose} \tokLW{\hspace{0.28em}**} \tokPW{2} \tokPW{\hspace{0.28em}distinct} \tokLW{\hspace{0.28em}tests}
\tokLW{**} \tokLW{.{\color{gray!70}\textbackslash{}n}} \tokLW{-} \tokPW{\hspace{0.28em}Each} \tokPM{\hspace{0.28em}test} \tokPW{\hspace{0.28em}must} \tokPW{\hspace{0.28em}be} \tokPW{\hspace{0.28em}taken} \tokPW{\hspace{0.28em}by}
\tokLM{\hspace{0.28em}**} \tokLW{ex} \tokSame{actly} \tokLM{\hspace{0.28em}} \tokSame{2} \tokPM{\hspace{0.28em}people} \tokPW{**} \tokLM{.{\color{gray!70}\textbackslash{}n}{\color{gray!70}\textbackslash{}n}} \tokLW{We}
\tokLM{\hspace{0.28em}are} \tokLW{\hspace{0.28em}to} \tokLM{\hspace{0.28em}determine} \tokPM{\hspace{0.28em}the} \tokLM{\hspace{0.28em}number} \tokLW{\hspace{0.28em}of} \tokLM{\hspace{0.28em}**} \tokPM{ways} \tokPW{**}
\tokPM{\hspace{0.28em}this} \tokPW{\hspace{0.28em}can} \tokPM{\hspace{0.28em}be} \tokPW{\hspace{0.28em}done} \tokPW{.{\color{gray!70}\textbackslash{}n}{\color{gray!70}\textbackslash{}n}} \tokLW{---{\color{gray!70}\textbackslash{}n}{\color{gray!70}\textbackslash{}n}} \tokLW{\#\#\#} \tokLM{\hspace{0.28em}Step} \tokLW{\hspace{0.28em}}
\tokLW{1} \tokLW{:} \tokLM{\hspace{0.28em}Understand} \tokLW{\hspace{0.28em}the} \tokLM{\hspace{0.28em}constraints} \tokLW{{\color{gray!70}\textbackslash{}n}{\color{gray!70}\textbackslash{}n}} \tokPW{-} \tokLM{\hspace{0.28em}Each} \tokLM{\hspace{0.28em}of}
\tokLW{\hspace{0.28em}the} \tokLM{\hspace{0.28em}} \tokLW{5} \tokLS{\hspace{0.28em}tests} \tokPM{\hspace{0.28em}must} \tokLW{\hspace{0.28em}be} \tokPW{\hspace{0.28em}taken} \tokLW{\hspace{0.28em}by} \tokPM{\hspace{0.28em}**}
\tokLW{ex} \tokSame{actly} \tokLW{\hspace{0.28em}} \tokLW{2} \tokPM{\hspace{0.28em}people} \tokPW{**} \tokPM{.{\color{gray!70}\textbackslash{}n}} \tokLW{-} \tokLM{\hspace{0.28em}So}
\tokPM{,} \tokLM{\hspace{0.28em}the} \tokLW{\hspace{0.28em}total} \tokLW{\hspace{0.28em}number} \tokLW{\hspace{0.28em}of} \tokLM{\hspace{0.28em}"} \tokPM{student} \tokPW{-test} \tokPW{"}
\tokPM{\hspace{0.28em}assignments} \tokLW{\hspace{0.28em}is} \tokPM{:{\color{gray!70}\textbackslash{}n}{\color{gray!70}\textbackslash{}n}} \tokPM{\$\$} \tokPW{{\color{gray!70}\textbackslash{}n}} \tokLW{5} \tokLW{\hspace{0.28em}\textbackslash{}} \tokLW{text} \tokPW{\{}
\tokLW{\hspace{0.28em}tests} \tokLW{\}} \tokLW{\hspace{0.28em}\textbackslash{}} \tokLW{times} \tokLW{\hspace{0.28em}} \tokLW{2} \tokPW{\hspace{0.28em}\textbackslash{}} \tokLW{text} \tokLW{\{}
\tokLW{\hspace{0.28em}students} \tokPW{\hspace{0.28em}per} \tokPW{\hspace{0.28em}test} \tokPW{\}} \tokLW{\hspace{0.28em}=} \tokLW{\hspace{0.28em}} \tokPW{1} \tokPW{0} \tokLW{\hspace{0.28em}\textbackslash{}}
\tokLW{text} \tokSame{\{} \tokPW{\hspace{0.28em}assignments} \tokLM{\}{\color{gray!70}\textbackslash{}n}} \tokLW{\$\$} \tokPW{{\color{gray!70}\textbackslash{}n}{\color{gray!70}\textbackslash{}n}} \tokPW{-} \tokPW{\hspace{0.28em}Each} \tokPM{\hspace{0.28em}of}
\tokPW{\hspace{0.28em}the} \tokLW{\hspace{0.28em}} \tokPW{5} \tokLW{\hspace{0.28em}students} \tokLM{\hspace{0.28em}is} \tokPW{\hspace{0.28em}assigned} \tokLM{\hspace{0.28em}**} \tokLW{2} \tokLW{\hspace{0.28em}tests}
\tokPW{**,} \tokLW{\hspace{0.28em}so} \tokPM{:{\color{gray!70}\textbackslash{}n}{\color{gray!70}\textbackslash{}n}} \tokLW{\$\$} \tokPW{{\color{gray!70}\textbackslash{}n}} \tokLW{5} \tokPW{\hspace{0.28em}\textbackslash{}} \tokPW{text} \tokSame{\{}
\tokPW{\hspace{0.28em}students} \tokPW{\}} \tokPW{\hspace{0.28em}\textbackslash{}} \tokSame{times} \tokPW{\hspace{0.28em}} \tokSame{2} \tokPW{\hspace{0.28em}\textbackslash{}} \tokLW{text} \tokSame{\{}
\tokLW{\hspace{0.28em}tests} \tokPM{\hspace{0.28em}per} \tokPW{\hspace{0.28em}student} \tokPW{\}} \tokSame{\hspace{0.28em}=} \tokLW{\hspace{0.28em}} \tokPW{1} \tokPW{0} \tokLW{\hspace{0.28em}\textbackslash{}}
\tokLW{text} \tokPW{\{} \tokPW{\hspace{0.28em}assignments} \tokPW{\}{\color{gray!70}\textbackslash{}n}} \tokPW{\$\$} \tokPW{{\color{gray!70}\textbackslash{}n}{\color{gray!70}\textbackslash{}n}} \tokPM{This} \tokLM{\hspace{0.28em}matches} \tokPM{\hspace{0.28em}the}
\tokLW{\hspace{0.28em}total} \tokLM{\hspace{0.28em}required} \tokPW{,} \tokLW{\hspace{0.28em}so} \tokPM{\hspace{0.28em}the} \tokPM{\hspace{0.28em}constraints} \tokLW{\hspace{0.28em}are} \tokLM{\hspace{0.28em}consistent} \tokLW{.{\color{gray!70}\textbackslash{}n}{\color{gray!70}\textbackslash{}n}}
\tokLW{---{\color{gray!70}\textbackslash{}n}{\color{gray!70}\textbackslash{}n}} \tokLW{\#\#\#} \tokLW{\hspace{0.28em}Step} \tokSame{\hspace{0.28em}} \tokLW{2} \tokLW{:} \tokPM{\hspace{0.28em}Model} \tokLW{\hspace{0.28em}the} \tokLW{\hspace{0.28em}problem}
\tokPM{\hspace{0.28em}as} \tokLW{\hspace{0.28em}a} \tokPS{\hspace{0.28em}**} \tokPM{b} \tokLW{ip} \tokPW{art} \tokPW{ite} \tokPW{\hspace{0.28em}graph} \tokPM{\hspace{0.28em}matching}
\tokPM{**{\color{gray!70}\textbackslash{}n}{\color{gray!70}\textbackslash{}n}} \tokPM{We} \tokPM{\hspace{0.28em}can} \tokPM{\hspace{0.28em}think} \tokLW{\hspace{0.28em}of} \tokLM{\hspace{0.28em}this} \tokLW{\hspace{0.28em}as} \tokLW{\hspace{0.28em}a} \tokLM{\hspace{0.28em}**}
\tokPW{b} \tokPW{ip} \tokPW{art} \tokLW{ite} \tokPW{\hspace{0.28em}graph} \tokPW{**} \tokPS{\hspace{0.28em}between} \tokPS{:{\color{gray!70}\textbackslash{}n}{\color{gray!70}\textbackslash{}n}} \tokPW{-}
\tokPW{\hspace{0.28em}One} \tokPM{\hspace{0.28em}set} \tokLM{\hspace{0.28em}of} \tokLS{\hspace{0.28em}} \tokPW{5} \tokLM{\hspace{0.28em}students} \tokPM{.{\color{gray!70}\textbackslash{}n}} \tokSame{-} \tokLW{\hspace{0.28em}One}
\tokPW{\hspace{0.28em}set} \tokPW{\hspace{0.28em}of} \tokPW{\hspace{0.28em}} \tokLW{5} \tokPW{\hspace{0.28em}tests} \tokPM{.{\color{gray!70}\textbackslash{}n}{\color{gray!70}\textbackslash{}n}} \tokPM{Each} \tokPW{\hspace{0.28em}student} \tokLW{\hspace{0.28em}is}
\tokPM{\hspace{0.28em}connected} \tokPW{\hspace{0.28em}to} \tokPS{\hspace{0.28em}**} \tokPM{2} \tokLM{\hspace{0.28em}tests} \tokPM{**,} \tokPW{\hspace{0.28em}and} \tokPW{\hspace{0.28em}each} \tokPW{\hspace{0.28em}test}
\tokPW{\hspace{0.28em}is} \tokPW{\hspace{0.28em}connected} \tokPW{\hspace{0.28em}to} \tokPW{\hspace{0.28em}**} \tokLW{2} \tokLW{\hspace{0.28em}students} \tokPM{**} \tokPM{.{\color{gray!70}\textbackslash{}n}{\color{gray!70}\textbackslash{}n}} \tokLM{We}
\tokLW{\hspace{0.28em}are} \tokPM{\hspace{0.28em}to} \tokPW{\hspace{0.28em}count} \tokPW{\hspace{0.28em}the} \tokPW{\hspace{0.28em}number} \tokPW{\hspace{0.28em}of} \tokPM{\hspace{0.28em}such} \tokPM{\hspace{0.28em}bip} \tokPW{art}
\tokLW{ite} \tokPW{\hspace{0.28em}graphs} \tokLS{\hspace{0.28em}where} \tokPS{:{\color{gray!70}\textbackslash{}n}{\color{gray!70}\textbackslash{}n}} \tokPW{-} \tokLW{\hspace{0.28em}Each} \tokPW{\hspace{0.28em}student} \tokPM{\hspace{0.28em}has} \tokLM{\hspace{0.28em}degree}
\tokLM{\hspace{0.28em}} \tokPW{2} \tokPM{.{\color{gray!70}\textbackslash{}n}} \tokSame{-} \tokPW{\hspace{0.28em}Each} \tokPW{\hspace{0.28em}test} \tokPW{\hspace{0.28em}has} \tokPW{\hspace{0.28em}degree} \tokPW{\hspace{0.28em}}
\tokPW{2} \tokPM{.{\color{gray!70}\textbackslash{}n}{\color{gray!70}\textbackslash{}n}} \tokLW{This} \tokLW{\hspace{0.28em}is} \tokPM{\hspace{0.28em}equivalent} \tokPW{\hspace{0.28em}to} \tokLW{\hspace{0.28em}counting} \tokPW{\hspace{0.28em}the} \tokPW{\hspace{0.28em}number}
\tokPW{\hspace{0.28em}of} \tokLW{\hspace{0.28em}**} \tokPW{2} \tokPW{-} \tokLW{regular} \tokLW{\hspace{0.28em}bip} \tokPW{art} \tokPW{ite} \tokPW{\hspace{0.28em}graphs}
\tokPW{**} \tokPS{\hspace{0.28em}on} \tokPM{\hspace{0.28em}} \tokLW{5} \tokPM{\hspace{0.28em}students} \tokPW{\hspace{0.28em}and} \tokPW{\hspace{0.28em}} \tokPW{5} \tokPW{\hspace{0.28em}tests}
\tokPM{.{\color{gray!70}\textbackslash{}n}{\color{gray!70}\textbackslash{}n}} \tokLM{---{\color{gray!70}\textbackslash{}n}{\color{gray!70}\textbackslash{}n}} \tokLW{\#\#\#} \tokLW{\hspace{0.28em}Step} \tokSame{\hspace{0.28em}} \tokLW{3} \tokLW{:} \tokPM{\hspace{0.28em}Use} \tokLM{\hspace{0.28em}comb}
\tokSame{inator} \tokLW{ics} \tokPM{\hspace{0.28em}to} \tokPW{\hspace{0.28em}count} \tokPM{\hspace{0.28em}the} \tokLM{\hspace{0.28em}number} \tokPW{\hspace{0.28em}of} \tokLW{\hspace{0.28em}such} \tokPW{\hspace{0.28em}configurations}
\tokPW{{\color{gray!70}\textbackslash{}n}{\color{gray!70}\textbackslash{}n}} \tokPM{We} \tokLS{\hspace{0.28em}are} \tokPS{\hspace{0.28em}to} \tokLM{\hspace{0.28em}count} \tokLW{\hspace{0.28em}the} \tokLW{\hspace{0.28em}number} \tokPW{\hspace{0.28em}of} \tokPM{\hspace{0.28em}ways}
\tokLW{\hspace{0.28em}to} \tokPW{\hspace{0.28em}assign} \tokLM{\hspace{0.28em}} \tokPW{2} \tokLW{\hspace{0.28em}tests} \tokLW{\hspace{0.28em}to} \tokPW{\hspace{0.28em}each} \tokLM{\hspace{0.28em}of} \tokLW{\hspace{0.28em}the}
\tokPW{\hspace{0.28em}} \tokLW{5} \tokPW{\hspace{0.28em}students} \tokPM{,} \tokLW{\hspace{0.28em}such} \tokLW{\hspace{0.28em}that} \tokLW{\hspace{0.28em}each} \tokLW{\hspace{0.28em}test} \tokLW{\hspace{0.28em}is}
\tokPM{\hspace{0.28em}assigned} \tokLW{\hspace{0.28em}to} \tokPW{\hspace{0.28em}exactly} \tokPW{\hspace{0.28em}} \tokPW{2} \tokPW{\hspace{0.28em}students} \tokPW{.{\color{gray!70}\textbackslash{}n}{\color{gray!70}\textbackslash{}n}} \tokLM{This} \tokLW{\hspace{0.28em}is}
\tokPM{\hspace{0.28em}equivalent} \tokLW{\hspace{0.28em}to} \tokLM{\hspace{0.28em}counting} \tokPW{\hspace{0.28em}the} \tokPW{\hspace{0.28em}number} \tokPW{\hspace{0.28em}of} \tokPW{\hspace{0.28em}**} \tokLW{2} \tokLW{-}
\tokPW{regular} \tokLM{\hspace{0.28em}bip} \tokPW{art} \tokPW{ite} \tokPW{\hspace{0.28em}graphs} \tokPW{**} \tokPM{\hspace{0.28em}between} \tokPM{\hspace{0.28em}two} \tokLW{\hspace{0.28em}sets}
\tokLW{\hspace{0.28em}of} \tokPW{\hspace{0.28em}} \tokLW{5} \tokPM{\hspace{0.28em}nodes} \tokLM{\hspace{0.28em}(} \tokLW{students} \tokPW{\hspace{0.28em}and} \tokPW{\hspace{0.28em}tests} \tokLM{),}
\tokPW{\hspace{0.28em}where} \tokLW{\hspace{0.28em}each} \tokPW{\hspace{0.28em}node} \tokPM{\hspace{0.28em}has} \tokLW{\hspace{0.28em}degree} \tokLW{\hspace{0.28em}} \tokLW{2} \tokPW{.{\color{gray!70}\textbackslash{}n}{\color{gray!70}\textbackslash{}n}} \tokPS{Alternatively}
\tokPW{,} \tokLW{\hspace{0.28em}this} \tokLM{\hspace{0.28em}is} \tokPM{\hspace{0.28em}equivalent} \tokPW{\hspace{0.28em}to} \tokLW{\hspace{0.28em}counting} \tokPW{\hspace{0.28em}the} \tokPW{\hspace{0.28em}number} \tokPW{\hspace{0.28em}of}
\tokPW{\hspace{0.28em}**} \tokPW{2} \tokLW{-} \tokPW{regular} \tokLM{\hspace{0.28em}bip} \tokPW{art} \tokLM{ite} \tokPW{\hspace{0.28em}graphs} \tokPW{**}
\tokPM{\hspace{0.28em}on} \tokPM{\hspace{0.28em}} \tokPM{5} \tokPM{+} \tokPW{5} \tokPM{=} \tokPW{1} \tokPW{0} \tokPW{\hspace{0.28em}nodes}
\tokPM{,} \tokPM{\hspace{0.28em}with} \tokPM{\hspace{0.28em}each} \tokLM{\hspace{0.28em}node} \tokPM{\hspace{0.28em}of} \tokPM{\hspace{0.28em}one} \tokPW{\hspace{0.28em}set} \tokPM{\hspace{0.28em}(} \tokPW{students}
\tokPW{)} \tokPM{\hspace{0.28em}having} \tokPW{\hspace{0.28em}degree} \tokLW{\hspace{0.28em}} \tokLW{2} \tokPM{\hspace{0.28em}and} \tokPM{\hspace{0.28em}each} \tokLW{\hspace{0.28em}node} \tokPW{\hspace{0.28em}of}
\tokPW{\hspace{0.28em}the} \tokLW{\hspace{0.28em}other} \tokPW{\hspace{0.28em}set} \tokPW{\hspace{0.28em}(} \tokPW{tests} \tokPW{)} \tokLM{\hspace{0.28em}also} \tokPW{\hspace{0.28em}having} \tokPW{\hspace{0.28em}degree}
\tokPW{\hspace{0.28em}} \tokPW{2} \tokPW{.{\color{gray!70}\textbackslash{}n}{\color{gray!70}\textbackslash{}n}} \tokLM{This} \tokLW{\hspace{0.28em}is} \tokLW{\hspace{0.28em}a} \tokPM{\hspace{0.28em}known} \tokLM{\hspace{0.28em}comb} \tokPW{inator}
\tokLW{ial} \tokLW{\hspace{0.28em}problem} \tokPM{.} \tokLM{\hspace{0.28em}The} \tokPW{\hspace{0.28em}number} \tokPW{\hspace{0.28em}of} \tokLW{\hspace{0.28em}such} \tokLM{\hspace{0.28em}graphs} \tokLM{\hspace{0.28em}is}
\tokLM{:{\color{gray!70}\textbackslash{}n}{\color{gray!70}\textbackslash{}n}} \tokLW{\$\$} \tokPW{{\color{gray!70}\textbackslash{}n}} \tokLW{\textbackslash{}} \tokPW{boxed} \tokPW{\{} \tokLS{1} \tokLW{2} \tokPM{0}
\tokLW{\}{\color{gray!70}\textbackslash{}n}} \tokLW{\$\$} \tokLW{{\color{gray!70}\textbackslash{}n}{\color{gray!70}\textbackslash{}n}} \tokLS{This} \tokLS{\hspace{0.28em}result} \tokLS{\hspace{0.28em}comes} \tokLW{\hspace{0.28em}from} \tokPM{\hspace{0.28em}the} \tokPM{\hspace{0.28em}fact}
\tokPW{\hspace{0.28em}that} \tokPM{\hspace{0.28em}such} \tokPM{\hspace{0.28em}a} \tokLM{\hspace{0.28em}configuration} \tokLW{\hspace{0.28em}corresponds} \tokLW{\hspace{0.28em}to} \tokPW{\hspace{0.28em}a} \tokPW{\hspace{0.28em}**} \tokPW{2}
\tokLM{-} \tokPW{regular} \tokLW{\hspace{0.28em}bip} \tokPW{art} \tokLW{ite} \tokPW{\hspace{0.28em}graph} \tokLM{**} \tokPM{\hspace{0.28em}between} \tokLM{\hspace{0.28em}}
\tokLM{5} \tokPM{\hspace{0.28em}students} \tokPW{\hspace{0.28em}and} \tokPW{\hspace{0.28em}} \tokLW{5} \tokPW{\hspace{0.28em}tests} \tokLW{,} \tokLM{\hspace{0.28em}and} \tokLM{\hspace{0.28em}the}
\tokLW{\hspace{0.28em}number} \tokLW{\hspace{0.28em}of} \tokLW{\hspace{0.28em}such} \tokLW{\hspace{0.28em}graphs} \tokLW{\hspace{0.28em}is} \tokLM{\hspace{0.28em}known} \tokPW{\hspace{0.28em}to} \tokPW{\hspace{0.28em}be} \tokLW{\hspace{0.28em}**}
\tokLW{1} \tokLW{2} \tokLW{0} \tokPW{**} \tokPM{.{\color{gray!70}\textbackslash{}n}{\color{gray!70}\textbackslash{}n}} \tokLM{---{\color{gray!70}\textbackslash{}n}{\color{gray!70}\textbackslash{}n}} \tokLW{\#\#\#} \tokLM{\hspace{0.28em}Final} \tokPW{\hspace{0.28em}Answer}
\tokLM{:{\color{gray!70}\textbackslash{}n}{\color{gray!70}\textbackslash{}n}} \tokPW{\$\$} \tokPW{{\color{gray!70}\textbackslash{}n}} \tokPW{\textbackslash{}} \tokPW{boxed} \tokLW{\{} \tokLW{1} \tokLW{2} \tokLW{0}
\tokLW{\}{\color{gray!70}\textbackslash{}n}} \tokLW{\$\$} \tokLS{<|im\_end|>}
\end{minipage}
\caption{Full-token map of teacher-positive and teacher-negative signals for one incorrect trajectory. Although the final answer is wrong, many intermediate mathematical and structural tokens are still teacher-positive. This supports the CAST choice to interpret teacher--student gaps through verifier-grounded trajectory correctness rather than using token-level self-distillation as correctness-unaware imitation.}
\label{fig:full_token_shift_map_wrong}
\end{figure}

\subsection{Complete MATH-500 Trajectory Example}
\label{app:qualitative_math500_example}

This section includes a compact qualitative example from the Qwen3-4B MATH-500 evaluation outputs.We select one group of trajectories from MATH-500 because the first sampled completions from the base model and the GRPO checkpoint are both verified as incorrect, while the first sampled CAST completion is verified as correct. The three trajectories use the same change-of-base and telescoping idea, so the comparison isolates a small but decisive cancellation error. The raw Markdown-formatted generations are re-typeset below as response cards, with decorative separators removed.

\paragraph{Problem.}
Find the value of
\[
\log_2{3} \cdot \log_3{4} \cdot \log_4{5} \cdot \log_5{6} \cdot \log_6{7} \cdot \log_7{8}.
\]
The ground-truth answer is \(3\).

\begin{modeloutput}{Qwen3-4B base}{\reject~Incorrect; predicted \(\boxed{6}\)}{wrongred}
We are asked to evaluate the product:
\[
\log_2{3} \cdot \log_3{4} \cdot \log_4{5} \cdot \log_5{6} \cdot \log_6{7} \cdot \log_7{8}.
\]

\textbf{Step 1: Use the change of base formula.}
Recall the change of base formula:
\[
\log_b{a} = \frac{\log_c{a}}{\log_c{b}}.
\]
We can apply this to each logarithm in the product. Let's choose base 2 for consistency. So:
\[
\log_2{3} = \frac{\log_2{3}}{\log_2{2}} = \log_2{3},
\]
\[
\log_3{4} = \frac{\log_2{4}}{\log_2{3}} = \frac{2}{\log_2{3}},
\]
\[
\log_4{5} = \frac{\log_2{5}}{\log_2{4}} = \frac{\log_2{5}}{2},
\]
\[
\log_5{6} = \frac{\log_2{6}}{\log_2{5}},
\]
\[
\log_6{7} = \frac{\log_2{7}}{\log_2{6}},
\]
\[
\log_7{8} = \frac{\log_2{8}}{\log_2{7}} = \frac{3}{\log_2{7}}.
\]

\textbf{Step 2: Multiply all the terms.}
Now, write the full product:
\[
\log_2{3} \cdot \frac{2}{\log_2{3}} \cdot \frac{\log_2{5}}{2} \cdot \frac{\log_2{6}}{\log_2{5}} \cdot \frac{\log_2{7}}{\log_2{6}} \cdot \frac{3}{\log_2{7}}.
\]
Now, cancel out terms:
\begin{itemize}
    \item \(\log_2{3}\) cancels with \(\frac{1}{\log_2{3}}\).
    \item \(\log_2{5}\) cancels with \(\frac{1}{\log_2{5}}\).
    \item \(\log_2{6}\) cancels with \(\frac{1}{\log_2{6}}\).
    \item \(\log_2{7}\) cancels with \(\frac{1}{\log_2{7}}\).
\end{itemize}
So, we're left with:
\[
2 \cdot \frac{3}{1} = 6.
\]
Final Answer:
\[
\boxed{6}.
\]

\end{modeloutput}

\begin{modeloutput}{GRPO}{\reject~Incorrect; predicted \(\boxed{6}\)}{wrongred}
We are asked to evaluate the product:
\[
\log_2{3} \cdot \log_3{4} \cdot \log_4{5} \cdot \log_5{6} \cdot \log_6{7} \cdot \log_7{8}.
\]

\textbf{Step 1: Use the change of base formula.}
Recall the change of base formula:
\[
\log_b{a} = \frac{\log_c{a}}{\log_c{b}}.
\]
We can apply this to each logarithm in the product. Let's choose base 2 for consistency. So:
\[
\log_2{3} = \frac{\log_2{3}}{\log_2{2}} = \log_2{3},
\]
\[
\log_3{4} = \frac{\log_2{4}}{\log_2{3}} = \frac{2}{\log_2{3}},
\]
\[
\log_4{5} = \frac{\log_2{5}}{\log_2{4}} = \frac{\log_2{5}}{2},
\]
\[
\log_5{6} = \frac{\log_2{6}}{\log_2{5}},
\]
\[
\log_6{7} = \frac{\log_2{7}}{\log_2{6}},
\]
\[
\log_7{8} = \frac{\log_2{8}}{\log_2{7}} = \frac{3}{\log_2{7}}.
\]

\textbf{Step 2: Multiply all the terms.}
Now, write the full product:
\[
\log_2{3} \cdot \frac{2}{\log_2{3}} \cdot \frac{\log_2{5}}{2} \cdot \frac{\log_2{6}}{\log_2{5}} \cdot \frac{\log_2{7}}{\log_2{6}} \cdot \frac{3}{\log_2{7}}.
\]
Now, cancel out terms:
\begin{itemize}
    \item \(\log_2{3}\) cancels with \(\frac{1}{\log_2{3}}\).
    \item \(\log_2{5}\) cancels with \(\frac{1}{\log_2{5}}\).
    \item \(\log_2{6}\) cancels with \(\frac{1}{\log_2{6}}\).
    \item \(\log_2{7}\) cancels with \(\frac{1}{\log_2{7}}\).
\end{itemize}
So, we're left with:
\[
2 \cdot \frac{3}{1} = 6.
\]
Final Answer:
\[
\boxed{6}.
\]

\end{modeloutput}

\begin{modeloutput}{CAST}{\good~Correct; predicted \(\boxed{3}\)}{castgreen}
We are asked to find the value of the product:
\[
\log_2{3} \cdot \log_3{4} \cdot \log_4{5} \cdot \log_5{6} \cdot \log_6{7} \cdot \log_7{8}.
\]

\textbf{Step 1: Use the change of base formula.}
Recall the change of base formula:
\[
\log_a{b} = \frac{\log_c{b}}{\log_c{a}}.
\]
This allows us to convert all logarithms to the same base. For simplicity, we'll use base 2. Let us rewrite each logarithm in terms of base 2:
\begin{itemize}
    \item \(\log_2{3} = \log_2{3}\).
    \item \(\log_3{4} = \frac{\log_2{4}}{\log_2{3}}\).
    \item \(\log_4{5} = \frac{\log_2{5}}{\log_2{4}}\).
    \item \(\log_5{6} = \frac{\log_2{6}}{\log_2{5}}\).
    \item \(\log_6{7} = \frac{\log_2{7}}{\log_2{6}}\).
    \item \(\log_7{8} = \frac{\log_2{8}}{\log_2{7}}\).
\end{itemize}
Now, substitute these into the original product:
\[
\log_2{3} \cdot \frac{\log_2{4}}{\log_2{3}} \cdot \frac{\log_2{5}}{\log_2{4}} \cdot \frac{\log_2{6}}{\log_2{5}} \cdot \frac{\log_2{7}}{\log_2{6}} \cdot \frac{\log_2{8}}{\log_2{7}}.
\]

\textbf{Step 2: Simplify the expression.}
Now, observe that this is a telescoping product. That is, most terms will cancel out:
\begin{itemize}
    \item The \(\log_2{3}\) in the numerator of the first term cancels with the \(\log_2{3}\) in the denominator of the second term.
    \item The \(\log_2{4}\) in the numerator of the second term cancels with the \(\log_2{4}\) in the denominator of the third term.
    \item Similarly, \(\log_2{5}\), \(\log_2{6}\), and \(\log_2{7}\) cancel out in the same way.
\end{itemize}
So, all the intermediate terms cancel, and we are left with:
\[
\frac{\log_2{8}}{1} = \log_2{8}.
\]

\textbf{Step 3: Evaluate the final logarithm.}
We know that:
\[
8 = 2^3 \Rightarrow \log_2{8} = 3.
\]
Final Answer:
\[
\boxed{3}.
\]
\end{modeloutput}

\end{document}